\documentclass[journal]{IEEEtran}
\usepackage{amsmath,amsfonts}
\usepackage{algorithm}
\usepackage{array}
\usepackage{textcomp}
\usepackage{stfloats}
\usepackage{url}
\usepackage{verbatim}
\usepackage{graphicx}
\usepackage{cite}
\usepackage{graphicx}%
\usepackage{multirow}%
\usepackage{amsmath,amssymb,amsfonts}%
\usepackage{amsthm}%
\usepackage{mathrsfs}%
\usepackage{xcolor}%
\usepackage{textcomp}%
\usepackage{manyfoot}%
\usepackage{booktabs}%
\usepackage{algorithm}%
\usepackage{algorithmicx}%
\usepackage{algpseudocode}%
\usepackage{listings}%

\usepackage{nicefrac}       
\usepackage{epsfig}
\usepackage{subfigure}
\usepackage{url}
\usepackage{bm}
\usepackage{footmisc}
\usepackage{booktabs}       
\usepackage{color}
\usepackage{makecell}
\usepackage{hyperref}       
\usepackage{lineno}   
\usepackage{soul}
\usepackage{stfloats} 

\def\R{\mathbb{R}}

\begin{document}

\title{TRG-Net: An Interpretable and Controllable \\Rain Generator}

\author{Zhiqiang~Pang,
        Hong~Wang,
        Qi~Xie,
        Deyu~Meng,~\IEEEmembership{Member,~IEEE},
        Zongben~Xu
\thanks{Zhiqiang~Pang, Hong~Wang, Qi~Xie, Deyu~Meng and Zongben~Xu are with School of Mathematics and Statistics, Xi'an Jiaotong University, Shaanxi, P.R. China. E-mail: xjtupzq@gmail.com, hongwang9209@hotmail.com, \{xie.qi, dymeng, zbxu\}@mail.xjtu.edu.cn}
}

\markboth{Journal of \LaTeX\ Class Files,~Vol.~14, No.~8, August~2021}%
{Shell \MakeLowercase{\textit{et al.}}: A Sample Article Using IEEEtran.cls for IEEE Journals}


\maketitle

\begin{abstract}
Exploring and modeling rain generation mechanism is critical for augmenting paired data to ease training of rainy image processing models. 
Most of the conventional methods handle this task in an artificial physical rendering manner, through elaborately designing fundamental elements constituting rains. This kind of methods, however, are over-dependent on human subjectivity, which limits their adaptability to real rains. In contrast, recent  deep learning methods have achieved great success by training a neural network based generator from pre-collected rainy image data. However, current methods usually design the generator in a ``black-box'' manner, increasing the learning difficulty and data requirements.
To address these issues, this study proposes a novel deep learning based rain generator, which fully takes the physical generation mechanism underlying rains into consideration and well encodes the learning of the fundamental rain factors (i.e., shape, orientation, length, width and sparsity) explicitly into the deep network. Its significance lies in that the generator not only elaborately design essential elements of the rain to simulate expected rains, like conventional artificial strategies, but also finely adapt to complicated and diverse practical rainy images, like deep learning methods. By rationally adopting filter parameterization technique, the proposed rain generator is finely controllable with respect to rain factors and able to learn the distribution of these factors purely from data without the need for rain factor labels.
Our unpaired generation experiments demonstrate that the rain generated by the proposed rain generator is not only of higher quality, but also more effective for deraining and downstream tasks compared to current state-of-the-art rain generation methods. Besides, the paired data augmentation experiments, including both in-distribution and out-of-distribution (OOD), further validate the diversity of samples generated by our model for in-distribution deraining and OOD generalization tasks. Code is available at  \href{https://github.com/pzq-xjtu/TRG-Net}{https://github.com/pzq-xjtu/TRG-Net}.
\end{abstract}


\section{Introduction}\label{sec:introduction}

\IEEEPARstart{E}{xploring} and modeling rain generation mechanism is critical for augmenting paired data to ease the training of rainy image processing models. This is especially meaningful for current deep learning based image rain removal methods, where the effectiveness is highly dependent on paired rainy and rainless images \cite{wang2022survey,yang2020single,li2019single,wang2020single}. However, ideal data pairs are always hard to collect, particularly in the complicated and diverse rain scenarios in practice. Besides, the rain generation task is also of great potential value in revealing insightful characteristics underlying real rains, helpful in guiding sound policies for dealing with severe weather emergencies \cite{bossu2011rain,fu2017removing,yang2017deep}.

\begin{figure}[tbp]
\begin{center}
\hspace{-0mm}\includegraphics[width=1\linewidth]{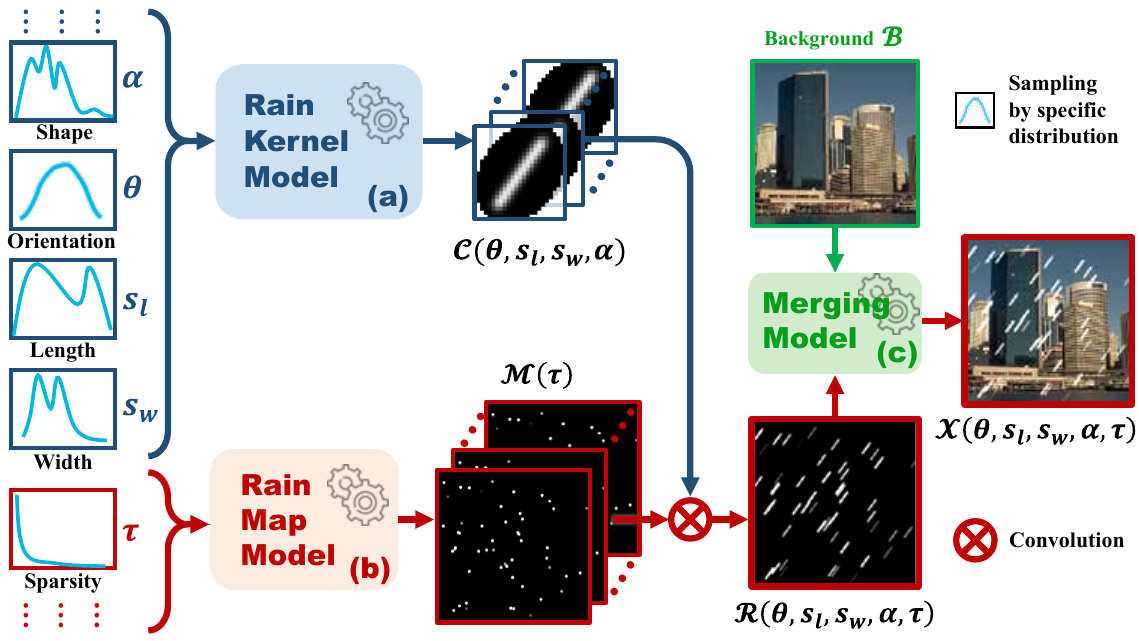}
\end{center}
 \vspace{-1mm}
  \caption{The pipeline for artificial rainy image synthesis based on physical rendering. It mainly contains three parts: (a) rain kernel model, (b) rain map model, and (c) merging model.}
\label{phy_quan}
\end{figure}

The early methods for investigating how to generate rainy images are mainly physical rendering based artificial synthesis manners \cite{garg2006photorealistic,garg2007vision,weber2015multiscale}. The physical rendering methods model rain streaks, the main rain appearance in rainy images, as the motion blur resulting from raindrops under the influence of gravity and wind, where the raindrops are characterised by an elaborate oscillation model. Specifically, as shown in Fig. \ref{phy_quan},  there are mainly three parts for generating a rainy image. The first part is the rain kernel model, for generating convolution kernels that involve the orientation, length, width and other shape information of the rain streaks. The second part is the rain map model, which is decided by the sparsity and position of the rains. By convolving the rain kernel and rain map, we can obtain a lifelike rain streak image. The final part is a merging model that 
{simulates the interaction between rain streaks and a background image (e.g., the fog-like rain formed by rain accumulation in 3D space).}
By carefully designing the three generation parts, rainy images can be synthesized with controllable shape, orientation, width, length and sparsity, even without any model training process. This manner has been widely used in the generation of rain scenes in movies, video games, animations and artistic works \cite{starik2003simulation,wang2006real}.

Although this category of methods can synthesize visually lifelike rainy images in the sense of human intuition, it still lacks realism to use the synthesized rainy/rainless images to train deep neural networks for deraining \cite{wei2019semi,yasarla2020syn2real}. This is because the rain factors (e.g., orientation, width, length and sparsity) of these methods are designed purely by human subjectivity, which inevitably leads to differences with the more complicated and diverse rains in real scenes.

Recently, deep learning (DL) based manners have been exploited for this task. The basic idea is to adopt generative adversarial networks (GANs) to train a rain generator and a discriminator, where the generator captures the rainy data distribution, and the discriminator distinguishes the difference between the real rains and the generated ones \cite{goodfellow2014generative,zhu2017unpaired,isola2017image}. In this way, the generated rains are forced to get close to the real rains in terms of the probability distribution through the training with real samples. It has been shown that the rainy/rainless image pairs generated in this way can help to achieve a great performance improvement in deraining tasks \cite{wang2021rain,choi2022synthesized,ni2021controlling,yu2023both}.

However, the current rain synthesis methods based on GANs still have obvious drawbacks. The most critical one is the design of rain generators. In particular, the current rain generators are usually assembled with some off-the-shelf network modules in commonly-adopted deep learning toolkits, e.g., convolutional neural networks (CNNs), which is indeed a ``black box'' without considering the intrinsic rain generation mechanism, and thus increases the learning difficulty and data requirements. Moreover, this insufficient knowledge modeling issue makes these methods hardly be effectively trained under the condition of inadequate paired training data, like those with unpaired rainy/rainless images \cite{wang2021rain,choi2022synthesized,ni2021controlling,wei2021deraincyclegan,chen2022unpaired}. 
It is thus critical to develop a rain generator capable of more faithfully delivering real rain generation mechanism for rain generation, with higher generation capability and lower data requirements. 

Fortunately, the generation mechanism of rain streaks has been well studied in the previous physical rendering based artificial synthesis methods, where one can find that rain streaks are mainly decided by their shapes, orientations, widths, lengths and sparsity \cite{garg2006photorealistic,garg2007vision,hu2019depth}, as shown in Fig. \ref{phy_quan}. 
Significant progress has been made in a very recent study \cite{yu2023both}, which incorporates these rain factors into deep networks and learns the connection between rain factors and deep networks using additional labels for rain factors. However, the requirement for these rain factor labels would further enhance the difficulty in data collecting.

In this paper, we focus on more intrinsically embedding the physical model of rain generation into deep networks to design a dedicated generator for rain generation, which can  be controlled with the rain factors and extract the distribution of rain factors without additional data labels on rain factors.
Specifically, we aim to mathematically model the relationship between rain factors (i.e., shape, orientation, width, length of the rain kernel and sparsity of the rain map) and  CNN network modules, while maintaining the learnable characteristics of CNN network. Since the calculations to rain factors are designed manually, there is thus no need for additional rain factor labels in the training process.
This is actually not trivial to achieve. For example, for controlling the orientation of rain streak generation, we have to dynamically rotate an entire CNN architecture in its spatial dimensions, which is not easy to implement with common CNN architecture. It is even more difficult to learn the orientation distribution of the rain streak in the whole rainy image dataset. 

To alleviate the above-mentioned difficulty, we exploit the filter parametrization technique and firstly design such an expected rain factor transformable convolution network for rain generation, where the aforementioned goals can all be finely achieved. The main contributions of this paper are summarized as follows:

\begin{figure}[tbp]
\begin{center}
\hspace{-2mm}\includegraphics[width=1\linewidth]{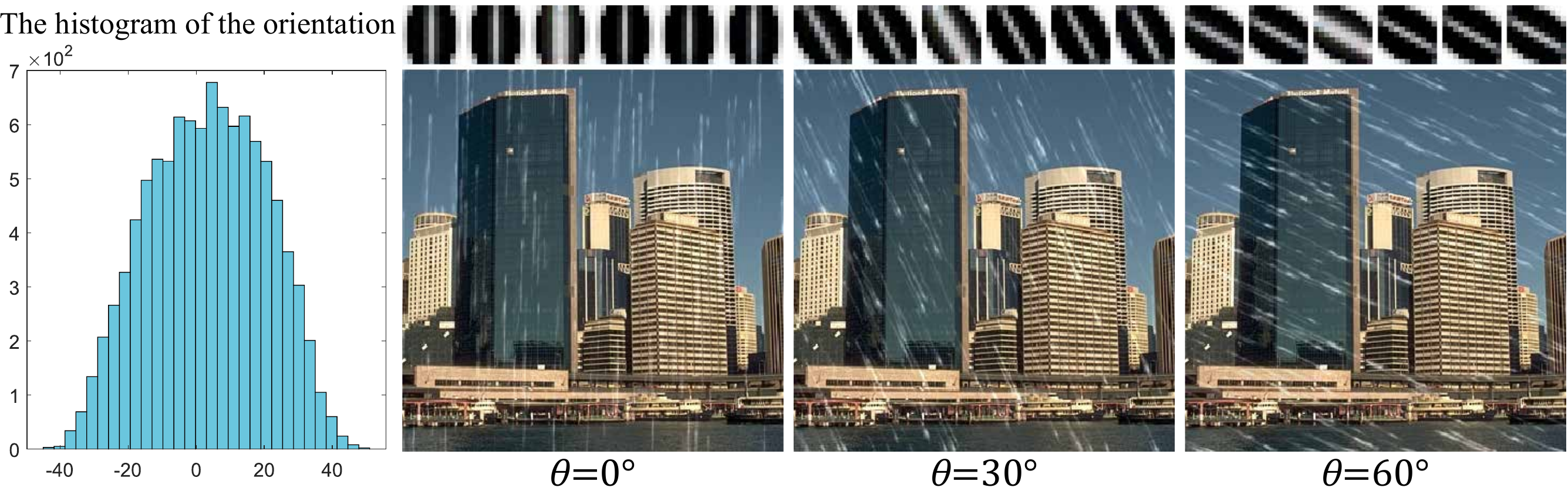}
\end{center}  
  \caption{The orientation distribution (left figure) extracted from Rain100L by the proposed TRG-Net, which enables to generate rains with required orientation by specifying the orientation degree $\theta$. Here, we generate the rains with orientations of $0^{\circ}$, $30^{\circ}$ and $60^{\circ}$, respectively.}
\label{Controllability_fig_theta}
\end{figure}

\begin{enumerate}
    \item We construct a new rain generator, which can properly extract fundamental rain factors underlying rainy images, including shape, orientation, width, length and sparsity, directly from data in a purely automatic manner. Its significance lies in that the model can both elaborately design fundamental elements to simulate expected rains like conventional artificial rendering manners, and finely adapt to complicated and diverse practical rain forms like recent deep learning methods. 
    (e.g., Fig. \ref{Controllability_fig_theta} shows the extracted distribution about the orientation factor from Rain100L, finely complying with its real situation). The model is thus with intrinsic interpretability and essential controllability, which can be trained without requiring data labels about the rain factors.
    \item To alleviate the difficulty in embedding controllable and learnable rain factors into deep networks, we construct a transformable convolution framework. Specifically, we exploit the filter parametrization technique \cite{xie2022fourier} for representing convolution filters to ameliorate the original discrete expression as continuous when constructing CNNs. This makes the expected transformation operators, such as rotation and scaling, able to be imposed on the convolution filters more easily and soundly, while keeping them learnable. Under such transformable convolutions, we can naturally construct  a transformable rain kernel model (in Sec. \ref{SecKernel}) and rotatable CNNs (in Secs. \ref{SecMap} and \ref{SecMerge}). Moreover, we present a new total variation (TV) regularizer, called rotatable TV regularizer (in Sec. \ref{lossfun}), capable of adaptively adjusting the orientation for calculating variations and adopting higher penalty along the rain streak orientation. The aforementioned functions could be potentially applied to a wider range of tasks.
    \item Our experiments show that the proposed model can effectively extract intrinsic physical mechanism implicitly contained in rainy images. The obtained rain factors, like orientation, scale and sparsity, are highly consistent with those reflected in the data, which can in turn be utilized to generate more variations of similar rain types with wider range of rain factors. 
    \item Comprehensive experiments in both unpaired rain generation and paired rain augmentation further substantiate the superiority of the proposed method beyond current state-of-the-art (SOTA) methods. Specifically, in the unpaired generation experiments, it is demonstrated that the rain generated by the proposed generator is not only of higher quality, but also more effective for deraining and downstream tasks (e.g., semantic segmentation), as compared with those generated by competing methods.  
    In the data augmentation experiments, including both in-distribution and out-of-distribution scenarios, it is further validated that the proposed method outperforms competing methods.
\end{enumerate}

The paper is organized as follows. Sec. \ref{Related Work} reviews necessary related works. Sec. \ref{Trans. Conv. Fram.} presents the transformable convolution framework for alleviating the difficulty that encodes controllable and learnable rain factors into deep networks. Sec. \ref{TRGNet} then proposes details of our proposed transformable rain generator. Sec. \ref{Experiments} demonstrates comprehensive experiments to substantiate the effectiveness of the proposed method. The paper is finally concluded with future work.


\section{Related Work} \label{Related Work}

\subsection{Rain Synthesis Methods}

The most straightforward way to collect rainy/rainless image pairs is to utilize rainy videos, where rainy images come from the video frames and rainless images are estimated from a sequence of video frames \cite{wang2019spatial,li2022toward,guo2023sky}. 
While this method is able to collect large-scale rainy images in real scenes, the resulting rainless images often contain remnants of rain streaks or lost image details.

Physical rendering based artificial synthesis is a common method that synthesizes rainy/rainless image pairs. The appearance of rain was studied early based on a raindrop oscillation model by Garg and Nayar \cite{garg2006photorealistic,garg2007vision}. The rain streaks, the main appearance of rain in rainy images, are considered to be the motion blur of the raindrops and can be synthesized with controllable shape, orientation, width, length and sparsity, by carefully setting the parameters of the rain model. Many 
large-scale rain datasets are constructed using the photorealistic rendering technique\footnote{https://www.photoshopessentials.com/photo-effects/rain/}, such as Rain100L \cite{yang2017deep}, Rain100H \cite{yang2017deep}, Cityscapes \cite{halder2019physics} and Kitti \cite{halder2019physics}. 
These large-scale paired rain datasets offer the training set for deraining networks and can quantitatively measure the effectiveness of deraining models. However, the rain streaks synthesized by these methods often tend to deviate from the real rains with much more complicated and diverse configurations. Such deviation usually leads to a degradation of the performance of the deraining models when dealing with the real rainy images.

Recently, GANs \cite{goodfellow2014generative,zhu2017unpaired} are employed for the rain generation where the probability distribution of the generated rain is forced to get close to that of the real rain through the training of certain generation network, which has achieved great success. Typically, 
\cite{wang2021rain,choi2022synthesized, ni2021controlling} generated rains using a generator stacked from some existing network modules and ignored the intrinsic rain generation mechanism, which is indeed a ``black box'' network. These methods tend to increase the learning difficulty and data requirements due to the neglecting of rain generation mechanism. Moreover, this insufficient knowledge modeling issue also makes these methods hardly able to be sufficiently trained on the condition of inadequate paired training data, like those with unpaired rainy/rainless images \cite{wang2021rain,choi2022synthesized}. 
Very recent work by Yu \emph{et al.} \cite{yu2023both} has been made significant progress in this task, which incorporates rain factors into deep networks using extra labels of these rain factors. However, the requirement for these rain factor labels would further enhance the difficulty in data collecting.


Different from previous rain synthesis methods, we focus on more intrinsically embedding the physical model of rain generation into deep networks to design a dedicated generator for rain generation, which can  be controlled with the rain factors and extract the distribution of rain factors without the need for additional data labels on rain factors.

\subsection{Single Image Rain Removal}
Since single image rain removal (SIRR) is highly related to the investigated rain generation task in this study, we also briefly review the research works along this research line. The SIRR task aims to reconstruct the rain-free image from an image degraded by rain streaks.
Recently, DL-based methods are the mainstream in SIRR. A flurry of network architectures, from simple CNNs \cite{fu2017removing,yang2017deep} to complicated architectures \cite{wang2019spatial,ren2019progressive,wang2021rcdnet} and transformer-based ones \cite{chen2023learning}, have been designed to handle this task. Although promising performance have been achieved, most of them are fully supervised and trained on synthetic rain data. The performance of these supervised derainers tend to drop dramatically when dealing with real-world rains, because there is a domain gap between the current synthesized and real rainy images.

There are some works trying to tackle this issue in a semi-supervised \cite{wei2019semi,yasarla2020syn2real} or unsupervised way \cite{ye2022unsupervised,chen2022unpaired}, but their performance is usually limited due to the lack of effective supervision information. Therefore, it is critical to design a rain generator for faithfully simulating real rain data with essentially more flexibly constructing paired datasets.

\section{Transformable Convolution Framework} \label{Trans. Conv. Fram.}

\begin{figure}
\begin{center}
\hspace{-2mm}\includegraphics[width=0.95\linewidth]{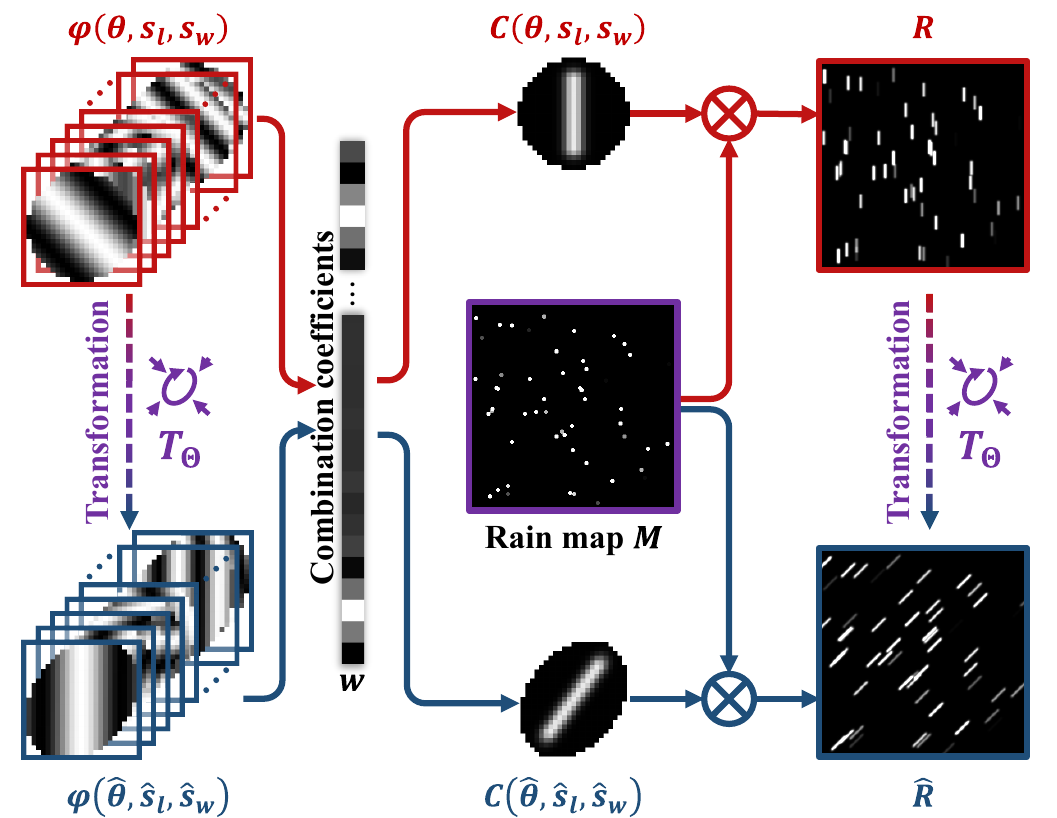}
\end{center}
\caption{Transform the convolution output by the transformation of the convolution filter, which is achieved by functional transforms on the underlying 2D basis functions. $\varphi(\theta,s_l,s_w)$ and $\varphi(\hat{\theta},\hat{s}_l,\hat{s}_w)$ are the basis function set (defined in Eqs. (\ref{basis set}) and (\ref{improved_Bases})) with different transformation parameters. $C(\theta,s_l,s_w)$ and 
$C(\hat{\theta},\hat{s}_l,\hat{s}_w)$ represent two convolution kernels (defined in Eq. (\ref{discretization})) that share the same combination 
coefficients $\bm{w}$, but have different transformation parameters. $R$ and $\hat{R}$ denote the convolution outputs of rain map $M$ with $C(\theta,s_l,s_w)$ and $C(\hat{\theta},\hat{s}_l,\hat{s}_w)$, respectively.
}
\label{controlability_convolution}
\end{figure}

In this section, we propose a transformable convolution framework to alleviate the difficulty in embedding controllable and learnable rain factors into deep networks. Traditional convolution kernels in deep networks are usually discrete, which can't be arbitrarily and accurately rotated or scaled. It is then natural to exploit continuous representation of convolution kernels for easily constructing transformable convolution. To this end, we employ filter parametrization methods to construct the overall framework of the transformable convolution,  as shown in  Fig. \ref{controlability_convolution}.

Specifically, the filter parametrization approach \cite{weiler2018learning} regards a discrete convolution kernel ${\phi}\in \mathbb{R}^{p \times p}$ as the discretization of a 2D continuous function $\tilde{\phi}$ on the $\left[-\nicefrac{(p-1)}{2},\nicefrac{(p-1)}{2}\right]^2$ area. 
A typical formulation for constructing $\tilde{\phi}(x)$ is:
\begin{equation}\label{Fourier}
\tilde{\phi}(x) = \sum_{n=1}^{N}w_{n}\varphi_{n}(x), \forall x \in \R^2,
\end{equation}
where
$\{w_n\}_{n=1}^{N}$ is representation coefficients, 
and $\varphi_{n}(x)$ denotes the $n^{th}$ basis function.

The key issue here is the choice of the basis function set. Especially, for rain generation task which requires pixel-level accuracy, it is important to choose a basis function set so that any  discrete filter can be accurately represented, i.e., for any ${\phi}$, there exists a set of $w_n$ so that $\tilde{\phi}(x_{ij}) =  {\phi}_{ij}$. Besides, another issue here is to avoid
aliasing effect\footnote{The aliasing effect here is caused by the insufficient sampling rate of discrete filter when the bases frequency is too high, resulting in an incorrect transformation result. For a detailed analysis of the aliasing effect please refer to \cite{xie2022fourier}.}, otherwise the kernel can be very unstable when being transformed \cite{xie2022fourier}.

Fortunately, it has been shown that Fourier series expansion based filter parametrization (FSE-FP) method \cite{xie2022fourier} can not only lead to zero representation error for any $\tilde{\phi}$, but also well release the aliasing effect, which finely meets the requirements of our task. Specifically, the basis function set proposed in FSE-FP method is
\begin{equation}\label{basis set}
  \Phi = \left\{\varphi_{kl}^c(x), \varphi_{kl}^s(x)|k,l =0,1,\cdots,p-1\right\},
\end{equation}
where
\begin{equation}\label{improved_Bases}
\begin{split}
  \varphi_{kl}^c(x)&\!=\!\Lambda(x)\cos\!\left(\frac{2\pi}{p}\!\left[k\!-\!\left\lfloor\frac{p}{2} \right\rfloor, l\!-\!\left\lfloor\frac{p}{2} \right\rfloor\right]\!\cdot\! x\right)\!,\\
  \varphi_{kl}^s(x)&\!=\!\Lambda(x)\sin\!\left(\frac{2\pi}{p}\!\left[k\!-\!\left\lfloor\frac{p}{2} \right\rfloor, l\!-\!\left\lfloor\frac{p}{2} \right\rfloor\right]\!\cdot\! x\right)\!,
\end{split}
\end{equation}
where $\Lambda(x)\geq 0$ is a radial mask function\footnote{The circular mask acts to limit the angle of basis functions, making them easier to transform. Please refer to \cite{xie2022fourier} for more details of $\Lambda(x)$.}, which satisfies $\Lambda(x) = 0$ if $\|x\|\geq(\nicefrac{p+1}{2})$ and $\lfloor{\cdot}\rfloor$ is the floor operator.

\textbf{Transformable convolution kernel.} Based on the parametrization method of Eqs. (\ref{Fourier}) and (\ref{improved_Bases}), we can implement any transformations on the convolution kernel $\tilde{\phi}(x)$ through the corresponding transformation of the basis functions, that is:
\begin{equation}\label{tranform_parameterization}
\begin{split}
\tilde{\phi}\left(x,\Omega\right) = \tilde{\phi}\left(T_{\Omega}\cdot x\right)
=\sum_{n=1}^{N}w_{n}\varphi_{n}\left(T_\Omega \cdot x\right),
\end{split}
\end{equation}
where $T_\Omega$ is the inverse transformation matrix with parameter set $\Omega$.
Then, the discretization of continuous transformable convolution kernel is
\begin{equation}\label{discretization}
  \left[{\phi}(\Omega)\right]_{ij} = \tilde{\phi}(x_{ij},\Omega) = \sum_{n=1}^{N}w_{n}\varphi_{n}\left(T_\Omega\cdot x_{ij}\right),
\end{equation}
It should be noted that $w_n, n = 1,\cdots,N$ are learnable parameters when utilizing this convolution kernel in the convolution layers of deep networks.

\textbf{Transformable convolution.} Since $\phi(\Omega)$ is a $p\times p$ matrix, which can be viewed as a discrete convolution kernel, we can exploit the common convolution between $\phi(\Omega)$ and any feature map to perform transformable convolution in the discrete domain. That is,
\begin{equation}\label{conv2d}
  F_{out}(\Omega) = F_{in}\otimes{\phi}(\Omega),
\end{equation}
where $F_{in}$ and $F_{out}$ represent the input and output feature maps, respectively, and $\otimes$ denotes the discrete convolution operator. For easy understanding, Fig. \ref{controlability_convolution} shows an example where the input is the rain map $M$ and the output is the controllable rain layer $R$.

Specifically, in this paper, we will need the rotation and scale transformations.
For rotation transformation only, we have $\Omega = \{\theta\}$, where $\theta$ denotes the rotation degree, and
\begin{equation}\label{Tmatrix1}
T_\Omega = T_{\{\theta\}}=  \begin{bmatrix} \cos{\theta} & -\sin{\theta}\\ \sin{\theta} & \cos{\theta} \end{bmatrix},
\end{equation}
When we need rotation and scale transformations simultaneously, we can set $\Omega = \left\{ \theta,s_l,s_w \right\}$, and
\begin{equation}\label{Tmatrix2}
T_\Omega = T_{\{\theta,s_l,s_w\}}=  \begin{bmatrix} s_w&0 \\ 0&s_l \end{bmatrix} \cdot \begin{bmatrix} \cos{\theta} & -\sin{\theta}\\ \sin{\theta} & \cos{\theta} \end{bmatrix},
\end{equation}
where $\theta,s_l,s_w$ are the rotation degree, the length parameter, and the width parameter, respectively.
It should be noted that  $s_l$ and $s_w$ are actually inversely proportional to the length and width, respectively, since $T_{\Omega}$ is an inverse transformation matrix.
In addition, we can find that all parameters (including transform parameters $\Omega$ and coefficient parameters $w_n$) of the convolution operation defined in Eq. (\ref{conv2d}) are easy to implement gradient backpropagation in deep learning frameworks. Therefore, the proposed transformable convolution should be potentially applicable to a wider range of the architecture designs in deep networks. In the following section, we employ the proposed transformable convolution framework to design a transformable rain generator that encodes the controllable and learnable rain factors.

\section{Transformable Rain Generator} \label{TRGNet}
The physical rendering based artificial synthesis method shown in Fig. \ref{phy_quan} is one of the most rational rainy image generation frameworks. However, it is difficult to formulate this framework into a learnable mathematical model. As an approximation, previous works \cite{li2018video,wang2020model} proposed a convolution sparse coding (CSC) based rain generation model:
\begin{equation}\label{CSC}
\mathcal{X} = \mathcal{R} +  \mathcal{B} =\tilde{\mathcal{C}}\otimes \tilde{\mathcal{M}}+\mathcal{B},
\end{equation}
where $\mathcal{X}$, $\mathcal{R}$, $\mathcal{B}\in\mathbb{R}^{H\times W\times 3}$ are RGB images, denoting the rainy image, the rain layer image, and the clean background image, respectively. $\tilde{\mathcal{C}}\in\mathbb{R}^{p\times p\times 3 \times K}$ is a rain kernel tensor,  representing $K$ rain kernels,   $\tilde{\mathcal{M}}\in\mathbb{R}^{H\times W\times K}$ denotes $K$ rain maps, $H$ and $W$ represent the height and width of the images, respectively, and $p$ denotes the spatial size of the rain kernel. Although $\tilde{\mathcal{C}}$ and $\tilde{\mathcal{M}}$ can be learned or generated by deep learning based manners, they are all fixed after the model is trained, which neglects essential dynamic rain factors like shape, orientation, length and so on. Therefore, the model (\ref{CSC}) is actually only suitable in deraining tasks, but not suitable for generating rainy images with diverse rain factors. Besides, this model simply adds the rain layer to the background image, which is also just a rough approximation to the real more complex situations.

In the following, we first introduce the proposed rain model, which fully takes necessary rain streak factors, including the shape, orientation, length, width and sparsity, into consideration. Then we present the proposed transformable rainy image generator, which takes the proposed rain model as the backbone. Lastly, we describe the training strategy and implementation details for the rainy image generator.

\subsection{Proposed Backbone Rain Model}\label{SecRainModel}
Although the artificial synthesis method shown in Fig. \ref{phy_quan} is difficult to be mathematically formulated with traditional convolution operators, it is actually not hard to be properly modeled with the proposed transformable convolution operator.
Following the pipeline of Fig. \ref{phy_quan}, we encode the shape, orientation, length, width factors of the rain into the transformable rain kernel, and encode the sparsity factor into the rain map. The expected rain model is expressed as:
\begin{equation} \label{rain_model}
\begin{split}
\!\mathcal{X}\!(\theta, s_l, s_w, \bm{\alpha}, \tau) &=\! \mbox{MerNet}\left(\mathcal{R}(\theta, s_l, s_w, \bm{\alpha},\tau), \mathcal{B}\right) \\
&= \!\mbox{MerNet}\!\left(\!\mathcal{C}(\theta, s_l, s_w, \bm{\alpha})\!\otimes\!  \mathcal{M}(\tau), \mathcal{B}\right),
\end{split}
\end{equation}
where  $\theta$, $s_l$, $s_w$, $\bm{\alpha}$ and $\tau$  represent the orientation, length, width, shape, and sparsity parameters, respectively. 
$\mathcal{X}(\theta, s_l, s_w, \bm{\alpha}, \tau)$, $\mathcal{R}(\theta, s_l, s_w, \bm{\alpha},\tau)$ and $\mathcal{B}$ denote the parameterized rainy image, the parameterized rain layer and the input clean background image, respectively.  
$\mathcal{C}(\theta, s_l, s_w,\bm{\alpha})\in\mathbb{R}^{p\times p\times 3 \times K}$ and $ \mathcal{M}(\tau)\in\mathbb{R}^{H\times W\times K}$ {are the rain kernel and the rain map, corresponding to $\tilde{\mathcal{C}}$ and $\tilde{\mathcal{M}}$ in Eq. (\ref{CSC}), respectively.}
 $\mbox{MerNet}(\cdot)$ is a merging network for merging the rain layer and background image, which simulates the interaction between rain and background image.

\begin{figure*}[ht]
\begin{center}
\hspace{-0mm}\includegraphics[width=1\linewidth]{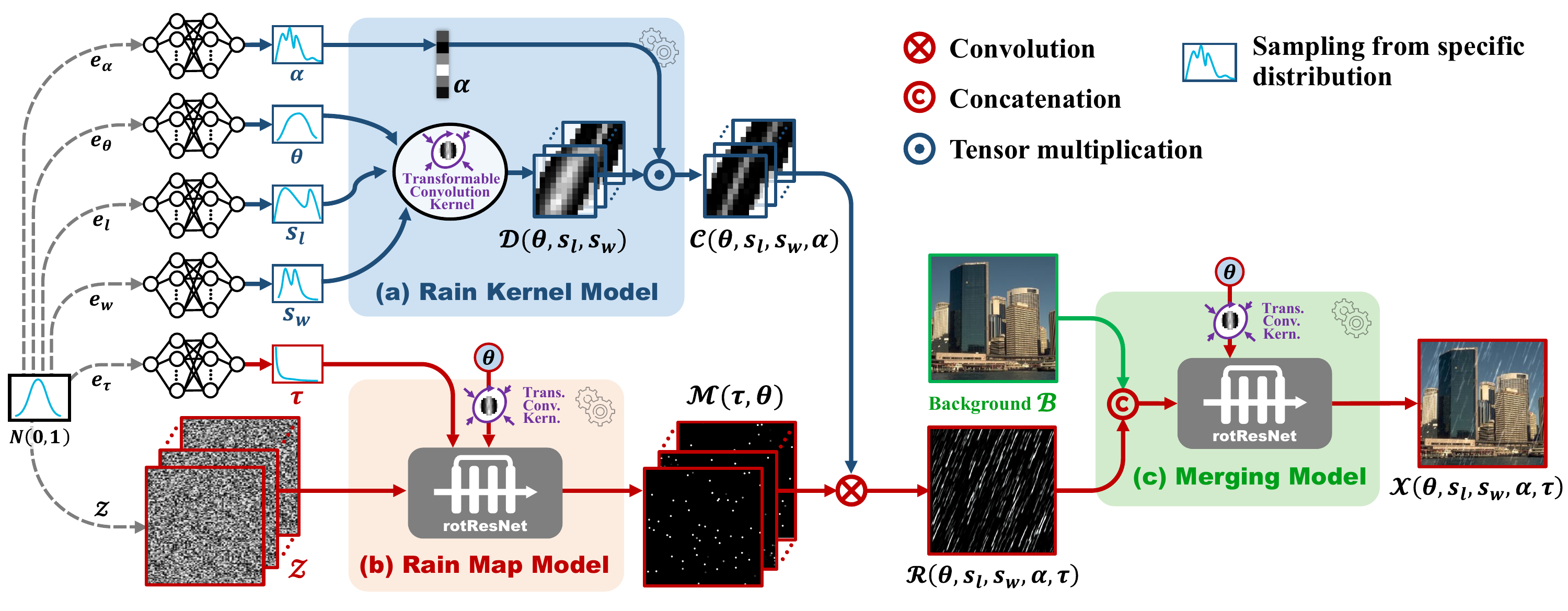}
\end{center}
  \caption{The proposed transformable rainy image generation network (TRG-Net), which takes the proposed rain model (\ref{rain_model}) as the backbone. Similar to the conventional artificial rainy image synthesis framework as shown in Fig. \ref{phy_quan}, it also consists of three parts: (a) the rain kernel model (Sec. \ref{SecKernel}), (b) the rain map model (Sec. \ref{SecMap}) and (c) the merging model (Sec. \ref{SecMerge}).
  }
\label{Schematic}
\end{figure*}

In Eq. (\ref{rain_model}), both the rain kernel $\mathcal{C}$ and the rain map $\mathcal{M}$ are parameterized with the rain factors. The parametrization of $\mathcal{C}$ is the main difficulty here, which is constructed based on the technique proposed in Eqs. (\ref{tranform_parameterization}), (\ref{discretization}) and (\ref{Tmatrix2}). 
While the rain map model is designed to incorporate sparsity as its main factor in controlling the presence of rain streaks, our empirical findings suggest that the orientation information of rain streaks is also certainly intertwined with the rain map. Therefore, it is necessary to consider the orientation information $\theta$ as an additional input to the rain map model. As a result, instead of using $\mathcal{M}(\tau)$, we use $\mathcal{M}(\tau, \theta)$ in this paper. 
Besides,  we utilize $\mbox{MerNet}(\cdot)$ to merge the rain layer and background image, which is also more rational than the simple plus operation in Eq. (\ref{CSC}), since the interaction between the rain layer and background in real rain is always more complex beyond simple plus. 
We can see that the model (\ref{rain_model}) actually takes (\ref{CSC}) as a special case, when we fix $\bm{\alpha}$, $\theta$, $s_l$, $s_w$, $\tau$ and set the merge model simply as plus operation.

\subsection{Transformable Rainy Image Generator}
Taking the proposed rain model (\ref{rain_model}) as a backbone, we further construct a transformable rainy image generation network (TRG-Net). The overall architecture of the proposed TRG-Net is shown in Fig. \ref{Schematic}, which mainly consists of a rain kernel model that generates the rain kernel $\mathcal{C}(\theta, s_l, s_w, \bm{\alpha})$, a rain map model to generate the rain map $\mathcal{M}(\tau, \theta)$, and a merging model for blending the rain layer image and the rainless background image. The construction details of these models are presented in the following sections.

\subsubsection{Rain Kernel Generation}\label{SecKernel}
\textbf{Rain kernel model.} 
We aim to design a new rain kernel model, where the distributions of rain factors can be learned from the data samples, rather than being manually preset as in conventional artificial rain synthesis. The key issue would lie in parameterizing these factors in a learnable way. Specifically, as shown in Fig. \ref{Schematic} (a), the considered factors for rain kernel include orientation degree, length, width and shape parameters, i.e, $\theta$, $s_l$, $s_w$ and $\bm{\alpha}$ in Eq. (\ref{rain_model}), where we utilize the proposed transformable convolution for parameterizing  orientation degree, length and width,  and use the  rain kernel dictionary representing technique \cite{wang2021rcdnet} 
for parameterizing the rain shapes. Formally, we propose the following rain kernel model:
\begin{equation}
\label{Eq:dynamic_kernel}
\begin{split}
\mathcal{C}(\theta, s_l, s_w, \bm{\alpha}) = \bm{\alpha} \odot \mathcal{D}(\theta, s_l, s_w),
\end{split}
\end{equation}
where $\mathcal{D}(\theta, s_l, s_w)\in \mathbb{R}^{p\times p\times 3 \times M}$ represents the rain kernel dictionary with $M$ elements.
 $\bm{\alpha}\in\mathbb{R}^{M\times K}$  denotes the $M$-dimensional coefficient vectors for the $K$ rain kernels,
 $\odot$ is the tensor multiplication along the $4^{th}$ dimension\footnote{For $\mathcal{D}\in \mathbb{R}^{p\times p\times 3 \times M}$ and $\bm{\alpha}\in\mathbb{R}^{M\times K}$, $(\mathcal{D} \odot\bm{\alpha})\in \mathbb{R}^{p\times p\times 3\times K}$ and
 $(\mathcal{D} \odot\bm{\alpha})_{\{:,:,:,k\}} = \sum_{m = 1}^{M}\mathcal{D}_{\{:,:,:,m\}}\bm{\alpha}_{mk}$.}. Besides, $\mathcal{D}(\theta, s_l, s_w)$ is parameterized in the manner of Eq. (\ref{discretization}). By substituting Eqs. (\ref{discretization}) and (\ref{Tmatrix2}), we can obtain that for any $i,j  = 1,2,\cdots,p$,  $c = 1,2,3$ and $k = 1,2,\cdots,K$,
\begin{equation}
\label{RainModel}
\begin{split}
\!\left[\mathcal{C}(\theta,\! s_l,\! s_w,\!\bm{\alpha})\right]_{ijck}\! =\!\sum_{mn}\! \bm{\alpha}_{mk} \!\cdot\! w_{mnc}^d\!\cdot\!\varphi_{n}\!\left(\!T_{\{\theta,s_l,s_w\}}\!\cdot\! x_{ij}\!\right)\!,
\end{split}
\end{equation}
where $\{\varphi_n\}_{n=1}^{N}$ are the basis functions defined in  Eqs. (\ref{basis set}) and (\ref{improved_Bases}), and the tensor $\bm{w}^d\in\mathbb{R}^{M\times N\times 3}$ denotes the to-be-learned coefficient parameters for the rain kernel dictionary.

\textbf{Rain kernel generator.} It is easy to find that the calculation about $\theta, s_l, s_w$ and $\bm{\alpha}$ only involves simple operations, all of which can be easily performed using popular deep learning tools, such as Pytorch, Tensorflow, and general gradient backpropagation algorithms, allowing us to learn their distribution automatically from data rather than manually set. Therefore, we can adopt lightweight fully connection networks (FCNs) for learning the distribution of these parameters, and train these FCNs with other parts of the proposed rainy image generator in an end-to-end way. Specifically, as shown in Fig. \ref{Schematic} (a),  we exploit the following four FCNs for rain factors generation\footnote{There is a reshape in output of the  FCN for $\bm{\alpha}$.}:
\begin{equation}
\label{Eq:para}\left\{
\begin{matrix}
\theta &=&\!\!\!\mbox{FCN}^{\theta}_{\bm{w}^\theta}(e_\theta), ~~e_\theta\sim{N(0,1)},\\
 s_{l} &=&\!\!\!\!\!\!\mbox{FCN}^{l}_{\bm{w}^l}(e_l), ~~e_l\sim{N(0,1)},\\
 s_{w} &=&\mbox{FCN}^{w}_{\bm{w}^w}(e_w), ~~e_w\sim{N(0,1)},\\
  \bm{\alpha} & =&\!\mbox{FCN}^{\alpha}_{\bm{w}^\alpha}(e_\alpha), ~~e_\alpha\sim{N(0,1)},
\end{matrix}\right.
\end{equation}
where all FCNs are input with Gaussian noise, and $\bm{w}^\theta$, $\bm{w}^l$, $\bm{w}^w$ and $\bm{w}^\alpha$ denote the to-be-learned parameters of the FCNs. Combining Eqs. (\ref{RainModel}) and (\ref{Eq:para}), we can obtain the entire rain kennel generator,
\begin{equation}\label{RainModelAll}
\begin{split}
  \mathcal{C}_{ijck}=&\sum_{mn}w_{mnc}^d\cdot \left[\mbox{FCN}^{\alpha}_{\bm{w}^\alpha}(e_\alpha)\right]_{mk}\cdot \\
  &~~\varphi_{n}\!\left(\!T_{\left\{\mbox{\scriptsize{FCN}}^{\theta}_{\bm{w}^\theta}(e_\theta),\mbox{\scriptsize{FCN}}^{l}_{\bm{w}^l}(e_l),\mbox{\scriptsize{FCN}}^{w}_{\bm{w}^w}(e_w)\right\}}\!\cdot\!x_{ij}\!\right),\!
\end{split}
\end{equation}
where $\mathcal{C}\in\mathbb{R}^{p\times p\times 3\times K}$ is the output rain kernel, and
$\{e_\theta, e_l, e_w, e_\alpha \}\sim N(0,1)$. One can refer to Fig. \ref{Schematic} (a) for an easy understanding of the rain kernel generator (\ref{RainModelAll}). For conciseness, we more briefly rewrite Eq. (\ref{RainModelAll}) as:
 \begin{equation}\label{CNet}
  \mathcal{C} =\mbox{KerNet}_{\bm{w}^c}(e_\theta, e_l, e_w, e_\alpha),
 \end{equation}
where $\bm{w}^c = \left\{\bm{w}^d, \bm{w}^\theta, \bm{w}^l, \bm{w}^w, \bm{w}^\alpha\right\}$ is all to-be-learnt parameters for rain kernel generation, which can be easily trained with off-the-shelf gradient backpropagation algorithms.

\subsubsection{Rain map Generation}\label{SecMap}
\textbf{Rain map model.}  The rain map $\mathcal{M}(\tau,\theta)$ mainly controls the position and sparsity of rain streaks. There are two key issues to be solved here: the first one is how to embed the orientation factor $\theta$ into the rain map model and make the rain map consistent with the rotation of the rain kernel; the second one is how to embed the sparsity parameter $\tau$ into the model. For the first issue, we introduce a rotatable ResNet (rotResNet), which is defined by replacing all the convolution kernels in the commonly used ResNet\cite{he2016deep} with the proposed rotatable convolution kernels defined in Eq. (\ref{discretization}). Formally, we have
\begin{equation}\label{rotResNet}
\mbox{rotResNet}_{\bm{w}^r}(\cdot, \theta) = \mbox{ResNet}_{\phi_{\bm{w}^r}(\theta)}(\cdot),
\end{equation}
where $\mbox{ResNet}_\phi(\cdot)$ denotes the commonly used ResNet whose convolution kernel is denoted by $\phi$, and $\phi_{\bm{w}^r}(\theta)$ is the parameterized convolution kernel defined by Eqs. (\ref{discretization}) and (\ref{Tmatrix1}), i.e., $\left(\phi_{\bm{w}^r}(\theta)\right)_{ij} = \sum_{n=1}^{N}w^r_{n}\varphi_{n}\left(T_{\{\theta\}}\cdot x_{ij}\right)$. $\mbox{rotResNet}_{\bm{w}^r}\left(\cdot,\theta\right)$ represents the rotatable ResNet with the rotation degree $\theta$ and the coefficient parameters $\bm{w}^r$.

For the second issue, a threshold ReLU operator is introduced for controlling the sparsity of rains. As a result, the proposed rain map model is
\begin{equation}
\label{Eq:rain map}
\begin{split}
\mathcal{M}(\tau,\theta) = \mbox{ReLU}(\mbox{rotResNet}_{\bm{w}^r}(\mathcal{Z}, \theta)-\tau),
 \end{split}
\end{equation}
where ReLU($\cdot$) denotes the ReLU activate function \cite{nair2010rectified}, and $\tau$ represents the sparsity factor of rain. 
{It's easy to see that the larger $\tau$ is, the more sparse the rain map will be.}
$\mathcal{Z}\in\mathbb{R}^{H\times W\times K}$ is the input random Gaussian noise. $\theta$ is the orientation degree, which is defined in Eq. (\ref{Eq:para}).

\textbf{Rain map generator.} Similar to Eq. (\ref{Eq:para}) for learning other rain factors, we adopt a lightweight FCN to learn the distribution of sparsity factor $\tau$:
\begin{equation}\label{tau}
  \tau = \mbox{FCN}^{\tau}_{\bm{w}^\tau}(e_\tau), ~~e_\tau\sim{N(0,1)},
\end{equation}
where $\bm{w}^\tau$ denotes the to-be-learned parameters. By substituting Eqs. (\ref{Eq:para}) (\ref{tau}) into (\ref{Eq:rain map}), we can obtain the entire rain map generator:
\begin{equation}
\label{RainMapGen}
\begin{split}
\!\mathcal{M}\!=\!\mbox{ReLU}\!\Big(\!\mbox{rotResNet}_{\bm{w}^r}\!\left(\!\mathcal{Z},\!\mbox{FCN}^{\theta}_{\bm{w}^\theta}\!(\!e_\theta\!)\!\right)\!-\!\mbox{FCN}^{\tau}_{\bm{w}^\tau}\!(\!e_\tau\!)\!\Big)\!,
 \end{split}
\end{equation}
where $\mathcal{M}\in\mathbb{R}^{H\times W\times K}$ is the output rain map, and
$\{\mathcal{Z}, e_\theta, e_\tau\}\sim N(0,1)$ are the input Gaussian noise. $\bm{w}^\theta$ and $e_\theta$ are shared with Eq. (\ref{CNet}). One can refer to Fig. \ref{Schematic} (b) for better understanding.  For simplicity, we rewrite Eq. (\ref{RainMapGen}) as:
\begin{equation}\label{MapNet}
  \mathcal{M} =\mbox{MapNet}_{\bm{w}^m}(\mathcal{Z},e_\tau,e_\theta),
\end{equation}
where  $\bm{w}^m = \left\{\bm{w}^r, \bm{w}^\tau, \bm{w}^\theta\right\}$ represents all the to-be-learned parameters. $\bm{w}^\theta$ and $e_\theta$ are shared with Eq. (\ref{CNet})

\subsubsection{Merging Rain Layer and Background}\label{SecMerge}
\textbf{Merging model.} For the merging network $\mbox{MerNet}(\cdot)$ in Eq. (\ref{rain_model}), we use a deep CNN for simulating the interaction between rain layer and the rainless background. Similar to the aforementioned rain map generator, we also adopt rotatable ResNet for the merging model, in order to be consistent with the rotation of the rain kernel. Formally, the merging model is

\begin{equation}\label{MerModel}
 \begin{split}
   \!\mathcal{X}\!(\!\theta,\! s_l,\! s_w,\! \bm{\alpha},\! \tau\!)\!=\! \mbox{rotResNet}_{\bm{w}^o}\!\left(\!\mbox{cat}(\!\mathcal{R}(\theta,\! s_l,\! s_w,\! \bm{\alpha},\! \tau\!),\! \mathcal{B}), \!\theta\!\right)\!,\! \\
 \end{split}
 \end{equation}
where $\text{rotResNet}_{\bm{w}^{o}}(\cdot,\theta)$ denotes the rotatable ResNet defined in Eq. (\ref{rotResNet}) with $\bm{w}^o$ denoting the to-be-learnt parameters. $\mbox{cat}(\cdot)$ represents the concatenation along the $3^{rd}$ dimension. $\mathcal{R}(\theta, \!s_l, \!s_w, \!\bm{\alpha}, \!\tau) = \mathcal{C}(\theta, s_l, s_w, \bm{\alpha})\otimes  \mathcal{M}(\tau,\theta)$ is the parameterized rain layer.
We can see that the model (\ref{MerModel}) is actually an executable version of the rainy image model (\ref{rain_model}).

\textbf{Merging network.} Combining model (\ref{MerModel}) with the FCN for $\theta$ in Eq. (\ref{Eq:para}), we can obtain the entire merging network:
\begin{equation}\label{MerModel2}
  \mathcal{X} = \mbox{rotResNet}_{\bm{w}^o}\!\left(\mbox{cat}(\mathcal{R}, \mathcal{B}), \mbox{FCN}^{\theta}_{\bm{w}^\theta}(e_\theta) \right)\!,
\end{equation}
where $\bm{w}^\theta$ and $e_{\theta}$ are shared with Eqs. (\ref{RainModelAll}) and (\ref{RainMapGen}). Please refer to Fig. \ref{Schematic} (c) for a better understanding of this design. We can then more concisely rewrite Eq. (\ref{MerModel2}) as:
\begin{equation}\label{MerNet}
  \mathcal{X} = \mbox{MerNet}_{\bm{w}^x}(\mathcal{R}, \mathcal{B}, e_\theta),
\end{equation}
where $\bm{w}^{x} = \{\bm{w}^o, \bm{w}^\theta\}$ denotes the to-be-learnt parameters.

\textbf{Rainy image generator.} By substituting Eqs. (\ref{CNet}), (\ref{MapNet}) and (\ref{MerNet}) into the backbone model (\ref{rain_model}), the proposed rainy image generator is
\begin{equation}\label{AllModel}
\begin{split}
  \mathcal{X} &=\mbox{TRG-Net}_{\bm{w}}\left(\mathcal{B}, \mathcal{Z}, \bm{e} \right)\\
  &\triangleq \mbox{MerNet}_{\bm{w}^x}\left(\mbox{KerNet}_{\bm{w}^c} (e_\theta, e_l, e_w, e_\alpha)\otimes\right.\\
   &~~~~~~~~~~~~~~~~~\left.\mbox{MapNet}_{\bm{w}^m}(\mathcal{Z},e_\tau,e_\theta), \mathcal{B}, e_\theta\right).
\end{split}
\end{equation}
where $\bm{w}= \{\bm{w}^x$, $\bm{w}^c$, $\bm{w}^m\}$ represents all the to-be-learnt parameters and $e=\{e_\theta, e_l, e_w, e_\alpha, e_\tau \}\sim N(0,1)$.
For brevity, we name this transformable rainy image generation network as TRG-Net.

\textbf{Remark.} It should be noted that the proposed framework is able to extract the distribution of rain factors purely from rain data without the need for rain factor labels using GAN-based manners due to these rain factors being intrinsically embedded. Besides, the proposed model possesses essential controllability. Specifically, after achieving $\bm{w}$ through learning from data, we can obtain all parameters.
Then either manually-set rain factors or FCN-generated rain factors can be utilized as the inputs to model (\ref{MerModel}). When using manually-set rain factors, the generated rainy images are essentially controlled. These merits are finely validated in Fig. \ref{Controllability_fig} below.

\subsection{Loss Function}\label{lossfun}
To capture the rain distribution in a real scene, the adversarial training strategy \cite{goodfellow2014generative} is applied to train TRG-Net. The basic loss function is:
\begin{equation} \label{loss}
\mathcal{L}_{adv}\left(\mbox{TRG-Net}_{\bm{w}}\left(\mathcal{B}, \mathcal{Z}, \bm{e} \right), {\mathcal{X}}_0 \right)+\lambda L_{rotTV},
\end{equation}
where $\mathcal{L}_{adv}$ is the GAN loss. ${\mathcal{X}}_0$  denotes the adversarial rainy image in training set. $\lambda$ is the trade-off parameter, and $L_{rotTV}$ represents specifically proposed regularizer for the task, which will be introduced in the following.

\begin{figure}[tbp]
\vspace{-0mm}
\begin{center}
\hspace{-0mm}\includegraphics[width=1\linewidth]{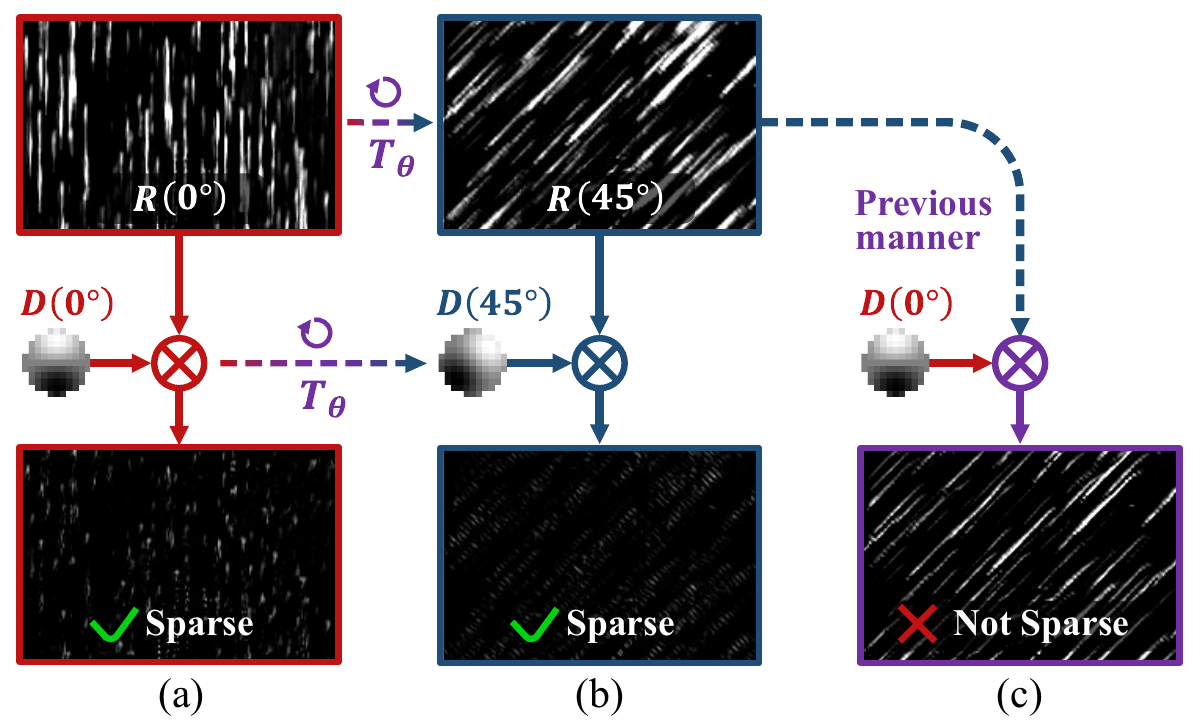}
\end{center}
  \caption{The differential field of a rain layer along its rain streak orientation is significantly sparser than those along other orientations. (a) The differential field of the rain $R(0^{\circ})$ with $0^{\circ}$ orientation along the $0^{\circ}$ orientation is sparse. (b) The differential field of the rain $R(45^{\circ})$ with $45^{\circ}$ orientation along the $45^{\circ}$ orientation is also sparse. (c) The differential field of the rain $R(45^{\circ})$ with $45^{\circ}$ orientation along the $0^{\circ}$ orientation is not sparse.
  }
\label{Fig:rotTV}
\end{figure}

\textbf{Rotatable Total Variation Regularizer.}
 Traditional total variation regularizer adopts the sparsity penalty on the first-order differential field of an image, which is in respect to the fact that most local parts of an image are smooth and edges in a natural image are sparse. Formally, TV regularizer can be presented as:
\begin{equation}\label{TV}
  L_{\text{TV}} = \|\hat{D}\otimes\hat{\mathcal{X}}\|_1,
\end{equation}
where $\hat{D}$ denotes the first-order difference filter and $\hat{\mathcal{X}}$ is an input image. 
It has been shown that TV regularizer is greatly helpful to improve the performance in many tasks \cite{rudin1992nonlinear,mahendran2015understanding,liu2021unpaired}.

However, the first order differential field of the  rain layer is generally very anisotropic, where the differential field along the rain streak orientation can be significantly sparser than those along other orientations \cite{zhuang2021reconciling,zhuang2022uconnet} (as shown in Fig. \ref{Fig:rotTV}). 
This fact yields the requirement of higher penalty along the rain streak orientation. Therefore, there is obvious room to improve the TV regularizer in rain layer regularization, since the difference filter $\hat{D}$ in the traditional TV regularizer can hardly be flexibly rotated to finely adapt different orientations of rain streaks in different images. 

Fortunately, in the proposed method, the orientation degree $\theta$ of the generated rain layer is able to be readily extracted by our model without extra effort. By adopting the transformable kernel  (\ref{discretization}) and the inverse rotation matrix (\ref{Tmatrix1}), the proposed rotated differential filter can be defined\footnote{The detail of derivation can be found in supplementary material.}. For $\forall i,j=1,2,\cdots,p$, we have
\begin{equation}\label{parD2}
  [D(\theta)]_{ij} = \frac{1}{p^2}\sum_{nst} \hat{D}_{st}\cdot\varphi_{n}(x_{st})\cdot \varphi_n(T_{\{\theta\}}\cdot x_{ij}), 
\end{equation}
where $\hat{D}$ is the vertical difference filter. $\theta$ denotes the rain orientation. Then the proposed rotatable total variation (rotTV) regularizer can be explicitly calculated by
\begin{equation}
L_{\text{rotTV}} = \left\| D(\theta) \otimes \mathcal{R} \right\|_1.
\end{equation}
In this paper, when we add this regularizer to the basic loss function, 
$\mathcal{R}$ is set as the generated rain layer (defined in Eq. (\ref{rain_model})), and $\theta = \mbox{FCN}^{\theta}_{\bm{w}^\theta}(e_\theta)$ and $\lambda$ is empirically set as 1. It is easy to see that the calculation that is caused by this term is readily backpropagated in deep networks.

\subsection{Implementation Details}

To generate the diverse, complex and long rain streaks in real world, we actually generate three rain kernels and convolute them with the rain maps to increase the receptive field of rain kernel.

Besides, it is easy to find that the rain kernels contain a clear physical meaning, which is quite different from other convolution kernels in deep networks. Therefore, we specifically design an initialization manner for them, achieving a streak-like initialization. Please refer to the supplemental material for more details for the rain kernels initialization and generation.

In order to adapt the rainy images to practical scenarios, where the orientations of rain streaks in a rainy image may be inconsistent, we can set $\theta$ as a vector with $M$ elements instead of a scalar (i.e., setting the output dimension number of $\mbox{FCN}_{\bm{w}^\theta}^\theta$ as $M$ instead of 1), where each  element in this orientation vector represents an individual orientation of  the element in the rain kernel dictionary. 

The number of rain kernel dictionary $M$ and the number of rain maps $K$ are set as $30$ and $6$, respectively. The size of rain kernels is $11\times 11$. 
For most of our experiments, we simply utilize 2 and 4 resblocks in the rain map model and the merging model, respectively. For datasets with heavy fog (i.e., Kitti \cite{tremblay2021rain}), a deeper merging model (with 10 resblocks) is employed to simulate the fog. Actually, more complex structures can be designed for these two parts based on actual needs in practice.


\section{Experimental Results} \label{Experiments}

In this section, we first validate the learning ability of the proposed rain generator and its controllability with respect to rain factors. Then, we comprehensively demonstrate the superiority of the proposed method in the sense of the quality of the generated samples, based on a series of experiments, including unpaired rain generation and paired rain augmentation, by comparison with the SOTA rainy image generators, e.g., VRG-Net \cite{wang2021rain}.

\subsection{Rain Generator Verification} \label{Dynamism and Controllability}

The proposed rain generator learns the rain factors by FCNs from data samples, which allows us to extract their distributions purely from the rain data set. We thus verify this capability on Rain100L \cite{yang2019joint}, a synthetic rain dataset containing 200 pairs of rainy and rainless images for training.

\textbf{Experiment settings.} To train TRG-Net by an adversarial strategy, following \cite{chen2022unpaired} and \cite{ye2022unsupervised}, we adopt a patch-based discriminator \cite{isola2017image}. Following \cite{wang2021rain}, the initial learning rates for the generator and the discriminator are $1\times 10^{-4}$ and $4\times 10^{-4}$, respectively. As suggested in \cite{gulrajani2017improved}, we update the discriminator five times for each generator updating. We utilize Adam algorithm to optimize the rain generation network. We train the proposed generator for 200 epochs with rotatable TV regularizer, and the network is trained for $3000$ iterations in each epoch. We randomly crop a $256\times 256$ patch as input in each iteration.

\begin{figure*}[htbp]
\vspace{-4mm}
\begin{center}
\hspace{-0mm}\includegraphics[width=1\linewidth]{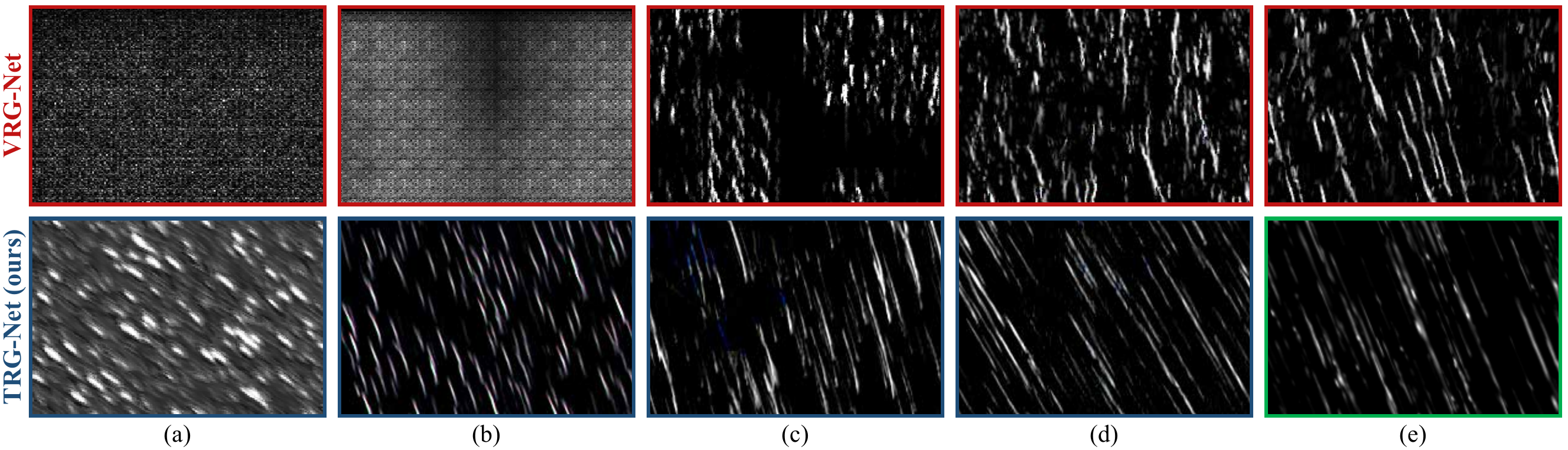}
\end{center}
\vspace{-2mm}
  \caption{The rain generation process on Rain100L in a paired training manner for both VRG-Net \cite{wang2021rain} and the proposed TRG-Net. (a)-(d): The rain generated by VRG-Net \cite{wang2021rain} (upper) and the proposed TRG-Net (lower) at random initialization, 1st, 10th and 60th epochs, respectively. (e) The rain generated by VRG-Net \cite{wang2021rain} at 700th epoch  (upper) and a typical reference rain in Rain100L \cite{yang2019joint} (lower).
  }
\label{Fig:rain_generation_epoch_main}
\vspace{-2mm}
\end{figure*}

\begin{figure*}[htbp]
\vspace{0mm}
\centering
\subfigure[The extracted orientation $\theta$ distribution from Rain100L.
The right four columns show the rainy images generated by inputing different orientation degree $\theta$ into TRG-Net.]
{
\includegraphics[width=1\linewidth]{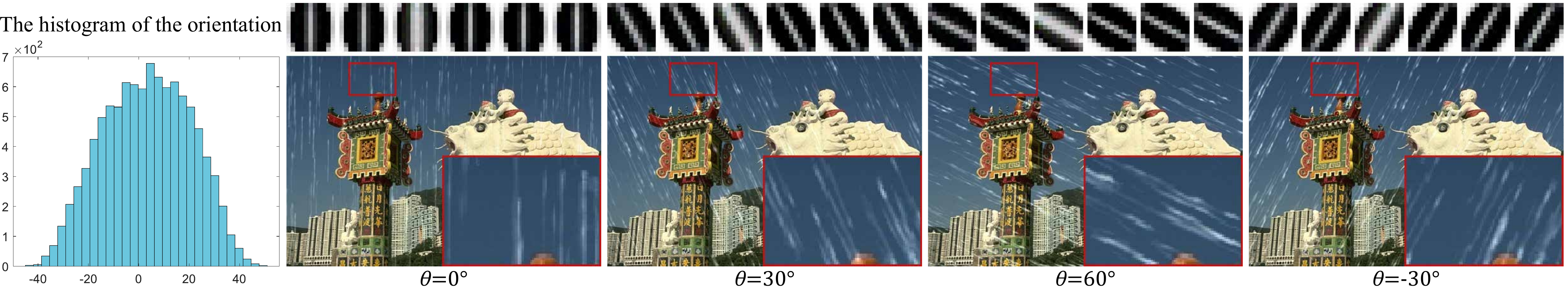}
}
\vspace{-2mm}
\centering
\subfigure[The learned distribution of width $s_w$ in Rain100L, whose values are all around $1.15$. 
 The right four columns show the rainy images generated by inputing different width parameters $s_w$ into TRG-Net.]
{
\includegraphics[width=1\linewidth]{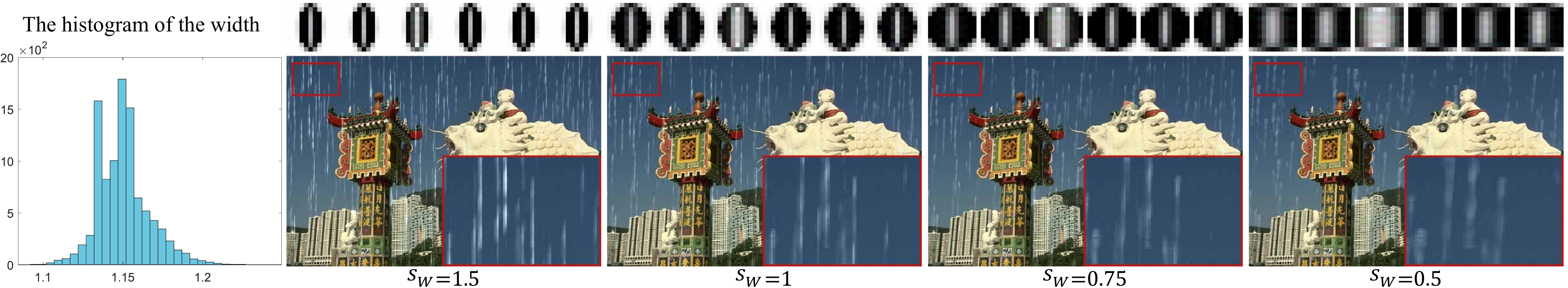}
}
\vspace{-2mm}
\centering
\subfigure[The learned length $s_l$ distribution from Rain100L, with values mostly between $0.6$ and $0.7$.
The right four columns show the rainy images generated by inputing different length parameter $s_l$ into TRG-Net.
]{
\includegraphics[width=1\linewidth]{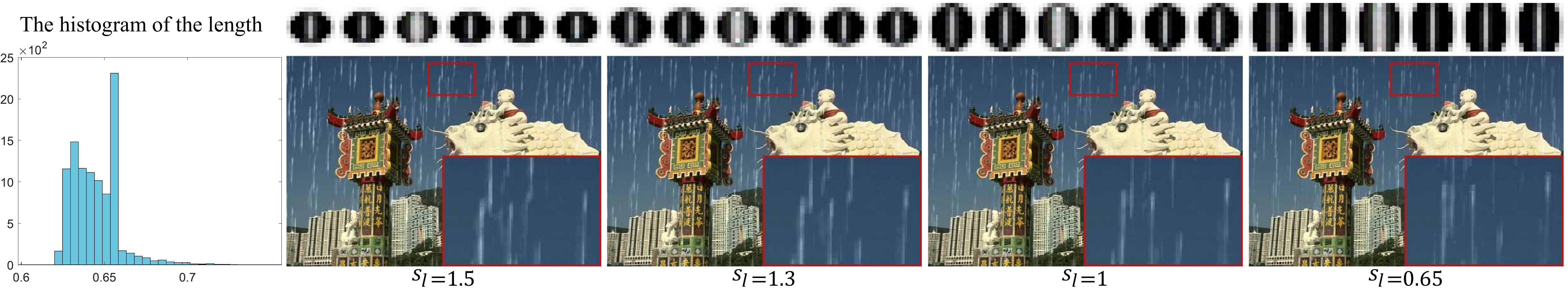}
}
\centering
\subfigure[The obtained sparsity $\tau$ distribution in Rain100L.
The right four columns show the rainy images generated by inputing different sparsity parameter $\tau$ into TRG-Net.
]{
\includegraphics[width=1\linewidth]{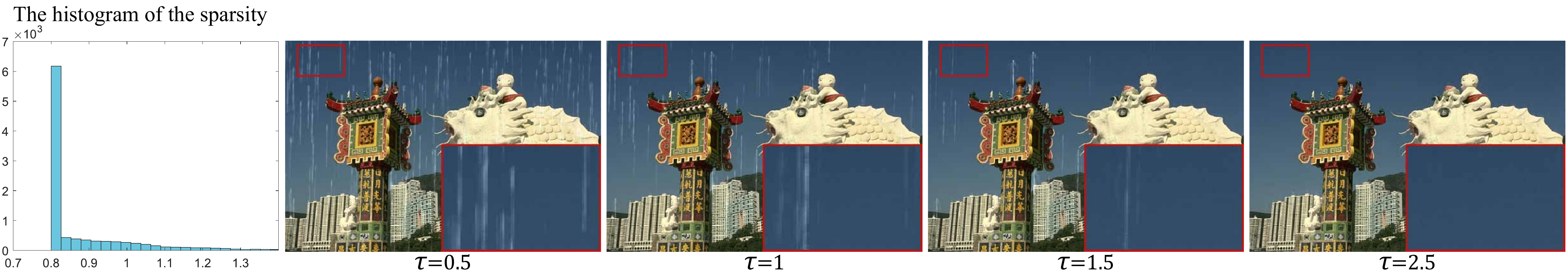}
}
\caption{The learned distributions of rain factors in Rain100L by the proposed TRG-Net, which enables to generate different types of rains by adjusting these rain factors in a controlled manner. (a)-(d) The first column are the learned distributions with respect to the orientation degree, width, length and sparsity, respectively, and the right four columns are some different rain types generated by adjusting these factors.}
\label{Controllability_fig}
\vspace{-4mm}
\end{figure*}

\textbf{Rain Generator Verification.} 
Fig. \ref{Fig:rain_generation_epoch_main} depicts the rain generation process in a paired training manner on Rain100L for both VRG-Net \cite{wang2021rain} and the proposed TRG-Net. One can see easily that the rain randomly initialized by TRG-Net already exhibits some rain patterns due to the embedding of a reasonable rain model, whereas the rain initialized by VRG-Net \cite{wang2021rain} appears to be mere noise. Furthermore, the rain generated by TRG-Net displays clear rain streaks even in the first epoch. These observations suggest that TRG-Net may exhibit relatively higher quality rain and a faster convergence rate for rain generation compared to VRG-Net, which will be further validated in subsequent experiments.

The learned distributions\footnote{The distribution is approximated by the histogram obtained by sampling 10000 times.} of rain factors in Rain100L are shown in Fig. \ref{Controllability_fig}, which can be easily used to further generate different rain types by adjusting rain factors in a controllable manner. As shown in Fig. \ref{Controllability_fig} (a), the learned orientation degree $\theta$ distribution is close to a Gaussian distribution, with values mostly between $-40$ and $40$, which complies with the intuitive orientation distribution in Rain100L \cite{bossu2011rain}. 
By model (\ref{MerModel}), we can manually input the orientation degree into the trained generator to control the orientation of the generated rain steaks. Fig. \ref{Controllability_fig} (a) shows the rain generated with input $\theta$ of $0^{\circ}$, $30^{\circ}$, $60^{\circ}$, $-30^{\circ}$, respectively.
It should be noted that this manner offers the capability to generate OOD rains, e.g., the $60^\circ$ rain in Fig. \ref{Controllability_fig} (a), which will be beneficial for deraining on OOD rain data sets (see OOD rain augmentation in Sec. \ref{Sec: ood aug.}). Yet this is difficult to be achieved by traditional DL-based generators, since black-box model can hardly generalize samples beyond the distribution of training data. Similarly, in Fig. \ref{Controllability_fig} (b), (c) and (d), we can observe that the proposed method can also learn the distributions of $s_l$, $s_w$ and $\tau$, respectively. We can control rain from thin to thick and from short to long by scaling the rain kernels according to different $s_w$ and $s_l$ values, respectively, and control the sparsity of rain by adjusting the input of $\tau$.

\subsection{Unpaired Rain Generation} \label{Unsupervised Rain Generation and Removal}

In this section, we first conduct unpaired rain generation on three synthetic rain datasets and a real rain dataset. Then deraining experiments on the pseudo-paired data generated by the rain generators are performed to evaluate the quality of the unpaired generated rainy images for deraining.

\subsubsection{Unpaired Rain Generation}

\begin{table}
 \caption{The FID and KID of unpaired rain generation on four rain datasets. \textbf{Bold} indicates the best result.}
  \label{Table:FID and KID}
 \centering  \setlength{\tabcolsep}{13pt}
    \begin{tabular}{lcccc}
    \toprule 
     & \multicolumn{2}{c}{VRG-Net \cite{wang2021rain}}  &  \multicolumn{2}{c}{TRG-Net}   \\
    \cmidrule(r){2-3}\cmidrule(r){4-5}
    Datasets &FID$\downarrow$ & KID$\downarrow$ &FID$\downarrow$ & KID$\downarrow$ \\
    \midrule
    Rain100L     &   61.18 & 0.0196   & \textbf{25.78} & \textbf{0.0047} \\
    Cityscapes   &   39.94 & 0.0306   & \textbf{20.34} & \textbf{0.0133} \\
    Kitti        &   56.98 & \textbf{0.0364}   & \textbf{51.44} & 0.0392 \\
    SPA-Data     &   68.34 & 0.0391   & \textbf{48.79} & \textbf{0.0204} \\
    \bottomrule
    \end{tabular}
  
  \vspace{-2mm}
\end{table}

\begin{figure}[!h]
\vspace{-0mm}
\hspace{-0mm}\includegraphics[width=1\linewidth]{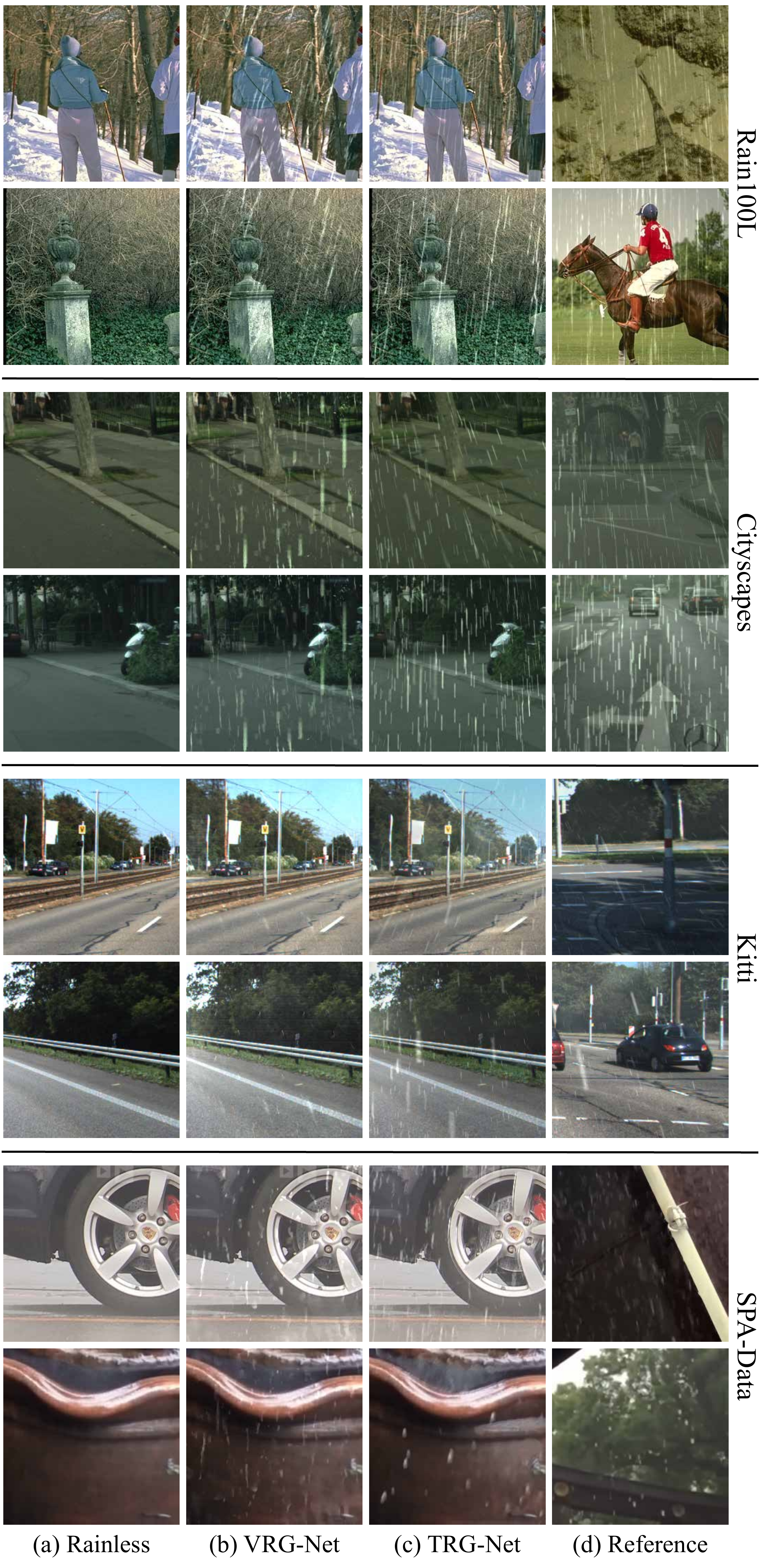}\centering
\vspace{-4mm}
  \caption{The results of unpaired rain generation by VRG-Net and the proposed TRG-Net on three synthetic datasets and a real dataset. (a)-(d) are the background images, the rainy images generated by VRG-Net, the rainy images generated by the proposed TRG-Net and the reference rainy images which are used for unpaired training, respectively. From top to bottom are Rain100L, Cityscapes, Kitti and SPA-Data datasets, respectively. Every dataset displays two generated samples.
  }
\label{Fig:unpaired_generated}
\vspace{-2mm}
\end{figure}

\textbf{Experiment settings.} We perform unpaired rain generation on three synthetic rain datasets, Rain100L, Cityscapes \cite{tremblay2021rain} and Kitti\footnote{
The rains in Cityscapes and Kitti are synthesised \cite{tremblay2021rain} with varying rainfall rate to evaluate the effect of rain on outdoor computer vision tasks. For them, we select the 100mm/hr rain as training set and test set in this experiment. As the rain orientations in Cityscapes and Kitti are correlated with the position, we have further included three learnable masks on the rain maps to distinguish different orientations in different positions. } \cite{tremblay2021rain}, and a real rain dataset, SPA-Data \cite{wang2019spatial}, to verify the superiority of our rain generator. Specifically, to construct a complete unpaired training dataset, we remove the rainless images of half of the data samples and remove the rainy images of the other half of the data samples. The rotatable TV regularizer is used for the training of Rain100L. {The generator is trained for 400 epochs on Cityscapes and Kitti, and 30 epochs and 250 epochs on Rain100L and SPA-Data, respectively.} The same patch-based discriminator \cite{isola2017image} is employed for all generators. Inspired by \cite{li2019heavy}, we use the high frequency components of the rainy images instead of the rainy image itself as input to the discriminator for better discrimination\footnote{Please refer to the supplementary material for more details}. The other training settings are the same as those used in Sec. \ref{Dynamism and Controllability}. We utilize Fr\'echet Inception Distance (FID) and Kernel Inception Distance (KID) as metrics for assessing the quality of generated rainy images, where lower values indicate better performance.

\begin{table*}[htbp]
  \caption{The quantitative results of all competing methods on synthetic and real datasets. A$^*$ indicates the  deraining results of PReNet \cite{ren2019progressive} trained on the pseudo-paired data generated by method A. The best result is highlighted with \textbf{bold}.
  }
  \label{Table: unpaired_deraining}
  \centering   \setlength{\tabcolsep}{15pt}
  {
  \begin{tabular}{lcccccccc}
    \toprule
    & \multicolumn{2}{c}{Rain100L}          &   \multicolumn{2}{c}{Cityscapes}  &  \multicolumn{2}{c}{Kitti}   &   \multicolumn{2}{c}{SPA-Data}      \\
    \cmidrule(r){2-3}\cmidrule(r){4-5}\cmidrule(r){6-7}\cmidrule(r){8-9}
    Methods    &   PSNR  & SSIM     & PSNR  & SSIM  & PSNR  & SSIM    & PSNR  & SSIM \\
    \midrule
    DSC \cite{luo2015removing}                      &   27.34 & 0.849   & 21.63 & 0.778 & 19.46 & 0.892  & 34.83 & 0.941  \\
    JCAS \cite{gu2017joint}                         &   28.54 & 0.852   & 22.63 & 0.877 & 18.92 & 0.867  & 34.95 & \textbf{0.945}           \\
    CycleGAN \cite{zhu2017unpaired}                 &   22.56 & 0.771   & 23.37 & 0.836 & 30.87 & 0.943  & 24.26 & 0.879 \\
    DerainCycleGAN \cite{wei2021deraincyclegan}     &   28.93 & 0.906   & 22.17 & 0.882 & 21.78 & 0.912  & 29.96 & 0.904 \\
    DCD-GAN \cite{chen2022unpaired}                 &   19.81 & 0.700   & 21.57 & 0.674 & 24.88 & 0.826  & 15.58 & 0.624 \\
    \midrule
    CycleGAN \cite{zhu2017unpaired}$^*$             &   25.11 & 0.832   & 22.62 & 0.908 & 26.88 & 0.937  & 22.71 & 0.866 \\
    DerainCycleGAN \cite{wei2021deraincyclegan}$^*$ &   28.50 & 0.910   & 21.10 & 0.871 & 22.31 & 0.938  & 22.01 & 0.863 \\
    DCD-GAN \cite{chen2022unpaired}$^*$             &   22.48 & 0.821   & 19.84 & 0.850 & 23.55 & 0.903  & 19.81 & 0.786 \\
    VRG-Net \cite{wang2021rain}$^*$                 &   27.84 & 0.855   & 21.94 & 0.768 & 21.08 & 0.906  & 34.81 & 0.931 \\
    TRG-Net$^*$                                     &   \textbf{32.33}  & \textbf{0.949}  & \textbf{24.86} & \textbf{0.959} & \textbf{32.15} & \textbf{0.983} & \textbf{35.80} & \textbf{0.945} \\
    \bottomrule
  \end{tabular}
  }
\end{table*}

\begin{figure*}[ht]
\vspace{-0mm}
\hspace{-0mm}\includegraphics[width=1\linewidth]{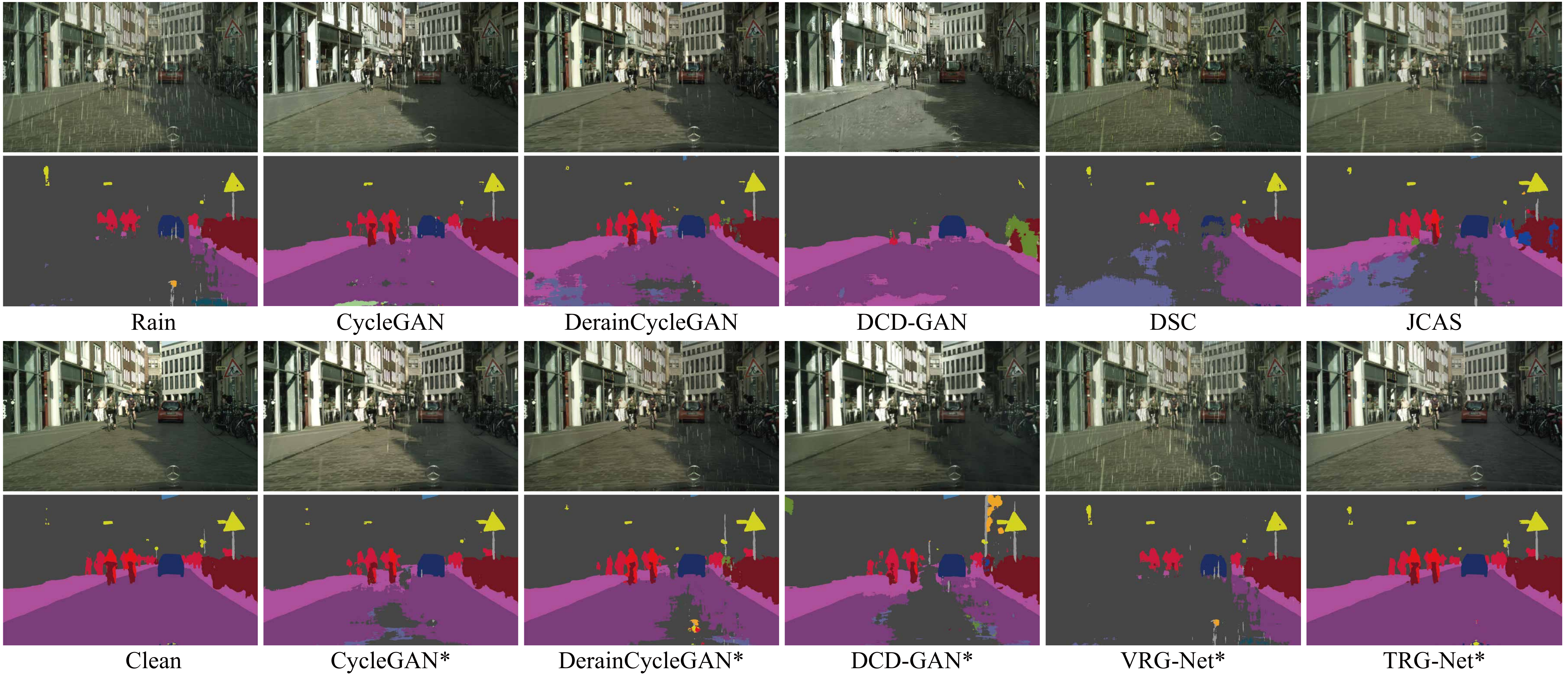}\centering
\vspace{-0mm}
  \caption{ The deraining results (the first row in each group) of all competing methods in an unpaired manner on Cityscapes,  and their corresponding semantic segmentation results (the second row in each group) on ERFNet \cite{romera2017erfnet}.
  }
\label{Fig:deraining_segment}
\vspace{-0mm}
\end{figure*}

\textbf{Experimental results.} Table \ref{Table:FID and KID} shows the FID and KID results on four datasets. Our model achieves lower FID and KID compared to VRG-Net on all datasets, expecting the KID on Kitti, which means that the rainy images generated by the proposed TRG-Net has a higher rain quality than VRG-Net \cite{wang2021rain} and is able to capture the rain generation mechanism under different environment conditions.  In Fig. \ref{Fig:unpaired_generated}, we show some typical results of the unpaired rain generation. It is easy to observe that the rain streaks generated by VRG-Net, which represents the CNN-based method, could be largely lacking in diversity, for example, the rain streaks in two random generation results on different background images are very similar, as shown in Fig. \ref{Fig:unpaired_generated} (b).  In contrast, the rainy images generated by the proposed TRG-Net not only better resemble to the original rainy images in their underlying rain patterns, i.e. the physical factors of rain such as shape and photometry are finely captured by our model, but also achieve more diversity in rain factors, such as orientation, length, width and sparsity. Actually, the diversity in rain factors of TRG-Net is easy to understand, and we can observe from Fig. \ref{Controllability_fig} that TRG-Net has extracted a specific distribution for rain factors, which certainly leads to the diversity in these factors. These results clearly substantiate the superiority of the proposed TRG-Net over the previous method.

\subsubsection{The Effectiveness of the Unpaired Generated Data for Deraining}

To demonstrate the effectiveness of the unpaired generated rainy images for deraining, we perform deraining experiments using the pseudo-paired data generated by the rain generators as the training data of deraining networks.

\textbf{Experiment settings.} We adopt PReNet \cite{ren2019progressive}, a simple deraining baseline, as the derainer to demonstrate the effectiveness of the generated pseudo train data sets for deraining. The training pairs for PReNet consist of rainless images from the train dataset of the unpaired rain generation experiment and their corresponding pseudo-paired rainy images generated by the rain generators. The competing unpaired methods include the model-based methods, DSC \cite{luo2015removing} and JCAS \cite{gu2017joint}, and the DL-based methods, CycleGAN \cite{zhu2017unpaired}, DerainCycleGAN \cite{wei2021deraincyclegan}and DCD-GAN \cite{chen2022unpaired}. Since the DL-based unpaired methods here are all CycleGAN-based framework which can also generate pseudo rainy images, we also train PReNet using the pseudo-paired data generated by these methods, for a fair comparison between these methods and rain generation methods. For convenience, we use the notation A$^*$ to denote the deraining results of PReNet trained on the pseudo paired data generated by method A. Peak-signal-to-noise (PSNR) and structure similarity (SSIM) are used to quantify the deraining performance, which are calculated in Y channel of YCbCr space following previous works \cite{chen2022unpaired,wang2021rain}.

\textbf{Experimental results.} Table \ref{Table: unpaired_deraining} lists the quantitative results of all competing methods on four datasets. We can observe that 
TRG-Net$^*$ outperforms all the other competing methods, which demonstrates the superiority of the unpaired generated rainy images by our model for deraining. Besides, to evaluate the effectiveness of unpaired deraining results for downstream tasks, we perform semantic segmentation experiments using ERFNet \cite{romera2017erfnet} on Cityscapes. 
In Fig. \ref{Fig:deraining_segment}, we provide the deraining results of all competing methods on a representative sample of Cityscapes and their corresponding semantic segmentation results\footnote{The quantitative results of semantic segmentation are presented in the supplementary material due to space limitations.}. It can be seen that TRG-Net$^*$ not only achieves clearly superior visual results in deraining, but also in segmentation.

\begin{table*}[htbp]
\caption{The deraining results on synthetic datasets. Baseline means the derainers trained on the original dataset without augmentation. VRG-Net and TRG-Net denote augmented training using VRG-Net and the proposed TRG-Net, respectively. 
\textbf{Bold} is the best result.
}
\label{Table-syn}
\centering \setlength{\tabcolsep}{10pt}
{
  \begin{tabular}{llcccccccccc}
    \toprule
    && \multicolumn{2}{c}{PReNet} & \multicolumn{2}{c}{SPANet} & \multicolumn{2}{c}{JORDER\_E} & \multicolumn{2}{c}{RCDNet} & \multicolumn{2}{c}{DRSformer}\\
    \cmidrule(r){3-4}\cmidrule(r){5-6}\cmidrule(r){7-8}\cmidrule(r){9-10}\cmidrule(r){11-12}
    Datasets&Methods&PSNR&SSIM&PSNR&SSIM&PSNR&SSIM&PSNR&SSIM&PSNR&SSIM \\
    \midrule
    \multirow{3}{*}{Rain100L}    
    & Baseline                   & 37.43 & 0.979 & 35.82 & 0.972 & 37.76 & 0.980 & 39.83 & 0.986 & 40.21 & \textbf{0.987} \\
    & VRG-Net & 37.79 & 0.980 & 36.07 & 0.973 & 38.38 & 0.981 & 40.05 & 0.986 & 40.39 & \textbf{0.987} \\
    & TRG-Net                    & \textbf{38.16} & \textbf{0.982} & \textbf{36.28} & \textbf{0.974} & \textbf{38.68} & \textbf{0.983} & \textbf{40.30} & \textbf{0.987} & \textbf{40.48} & \textbf{0.987} \\
    \midrule
    \multirow{3}{*}{Rain100L-S} 
    & Baseline                   & 36.42 & 0.974 & 35.01 & 0.967          & 36.47 & 0.973 & 39.08 & 0.984 & 39.09 & 0.983 \\
    & VRG-Net & 36.89 & 0.976 & 35.60 & \textbf{0.970} & 37.30 & 0.976 & 39.45 & 0.985 & 39.25 & 0.984 \\
    & TRG-Net                    & \textbf{37.21} & \textbf{0.978} & \textbf{35.70} & \textbf{0.970} & \textbf{37.50} & \textbf{0.978} &\textbf{39.63} & \textbf{0.986} & \textbf{39.89} & \textbf{0.986}\\
    \midrule
    \multirow{3}{*}{Rain100H} 
    & Baseline                   & 30.16 & 0.908 & 27.22 & 0.866 & 29.80 & 0.895 & 31.05 & 0.909 & 32.44 & \textbf{0.926} \\
    & VRG-Net & \textbf{30.19} & \textbf{0.909} & 27.30 & 0.866 & 30.20 & 0.900 & \textbf{31.16} & \textbf{0.911} & \textbf{32.46} & 0.924 \\
    & TRG-Net                       & {30.16} & {0.908} & \textbf{27.36} & \textbf{0.867} & \textbf{30.33} & \textbf{0.902} & {31.06} & \textbf{0.911} & \textbf{32.46} & 0.925 \\
    \midrule
    \multirow{3}{*}{Rain100H-S} 
    & Baseline                   & 27.83 & 0.878 & 26.53 & 0.854 & 27.32 & 0.864 & 29.52 & 0.896 & 29.62 & 0.897 \\
    & VRG-Net & 28.48 & 0.885 & 26.91 & 0.860 & 28.93 & 0.885 & 30.43 & 0.904 & 29.86 & 0.897 \\
    & TRG-Net                & \textbf{28.58} & \textbf{0.888} & \textbf{27.01} & \textbf{0.862} & \textbf{29.00} & \textbf{0.886} & \textbf{30.52} & \textbf{0.905} & \textbf{30.27} & \textbf{0.906} \\
    \bottomrule
  \end{tabular}
}
\vspace{-0mm}
\end{table*}

\begin{table*}[htbp]
\caption{The generalization performance on SPA-Data.
Baseline means the derainers trained on the original dataset without augmentation. VRG-Net and TRG-Net denote augmented training using VRG-Net and the proposed TRG-Net, respectively.  
The best result is highlighted with \textbf{bold}.}
\label{Table-spa}
\centering \setlength{\tabcolsep}{10pt}
{
  \begin{tabular}{llcccccccccc}
    \toprule
    && \multicolumn{2}{c}{PReNet} & \multicolumn{2}{c}{SPANet} & \multicolumn{2}{c}{JORDER\_E} & \multicolumn{2}{c}{RCDNet} & \multicolumn{2}{c}{DRSformer}\\
    \cmidrule(r){3-4}\cmidrule(r){5-6}\cmidrule(r){7-8}\cmidrule(r){9-10}\cmidrule(r){11-12}
    Datasets&Methods&PSNR&SSIM&PSNR&SSIM&PSNR&SSIM&PSNR&SSIM&PSNR&SSIM \\
    \midrule
    \multirow{3}{*}{Rain100L} 
    & Baseline                   & 34.95 & 0.941 & 35.18 & 0.946 & 35.02 & 0.942 & 34.88 & 0.938 & 34.48 & 0.947 \\
    & VRG-Net & 34.97 & 0.943 & 35.43 & 0.946 & 35.10 & 0.942 & 34.84 & 0.939 & \textbf{34.77} & 0.949 \\
    & TRG-Net                   & \textbf{35.50} & \textbf{0.949} & \textbf{35.50} & \textbf{0.948} & \textbf{35.78} & \textbf{0.950} & \textbf{35.59} & \textbf{0.948} & \textbf{34.77} & \textbf{0.952} \\
    \midrule
    \multirow{3}{*}{Rain100L-S} 
    & Baseline                   & 34.88 & 0.941 & 35.17 & 0.945 & 35.14 & 0.942 & 35.00 & 0.941 & 34.41 & 0.948 \\
    & VRG-Net & 35.08 & 0.943 & \textbf{35.36} & \textbf{0.947} & 35.17 & 0.943 & 35.03 & 0.941 & 34.52 & 0.948 \\
    & TRG-Net & \textbf{35.24} & \textbf{0.947} & 35.35 & \textbf{0.947} & \textbf{35.65} & \textbf{0.949} & \textbf{35.53} & \textbf{0.949} & \textbf{34.96} & \textbf{0.952} \\
    \midrule
    \multirow{3}{*}{Rain100H} 
    & Baseline                   & 33.10 & 0.934 & 32.72 & 0.942 & 33.59 & 0.941 & 34.25 & \textbf{0.944} & 34.25 & 0.945 \\
    & VRG-Net & {33.99} & \textbf{0.943} & 32.77 & 0.942 & 34.76 & \textbf{0.946} & {33.70} & \textbf{0.944} & 34.47 & 0.945 \\
    & TRG-Net       & \textbf{34.14} & \textbf{0.943} & \textbf{33.22} & \textbf{0.943} & \textbf{35.06} & \textbf{0.946} & \textbf{34.26} & \textbf{0.944} & \textbf{34.52} & \textbf{0.948} \\
    \midrule
    \multirow{3}{*}{Rain100H-S} 
    & Baseline                   & 33.73 & 0.940 & 31.50 & 0.931 & 33.24 & 0.935 & 34.36 & 0.944 & 34.22 & 0.946 \\
    & VRG-Net & 33.76 & 0.942 & 31.49 & 0.931 & 33.18 & 0.934 & 34.05 & 0.943 & 33.54 & 0.943 \\
    & TRG-Net       & \textbf{34.24} & \textbf{0.943} & \textbf{31.85} & \textbf{0.933} & \textbf{33.49} & \textbf{0.939} & \textbf{34.52} & \textbf{0.945} & \textbf{34.24} & \textbf{0.948} \\
    \bottomrule
  \end{tabular}
}
\end{table*}

\subsection{Rain Data Augmentation for Paired Data Set} \label{Rain Data Augmentation}



In this section, we perform rain data augmentation, including in-distribution and out-of-distribution augmentation, for deraining on paired datasets. Specifically, we first conduct in-distribution rain augmentation for deraining on synthetic datasets (Sec. \ref{Sec: syn. aug.}) and real SPA-Data (Sec. \ref{Sec: real aug.}) to demonstrate the diversity of rain generated by the proposed method. Then, OOD rain augmentation are implemented in Sec. \ref{Sec: ood aug.} to further validate the controllability superiority of TRG-Net.

\subsubsection{Rain Data Augmentation on Synthetic Datasets} \label{Sec: syn. aug.} 
\textbf{Datasets and deraining models.} Two commonly used synthetic rain datasets, Rain100L and Rain100H \cite{yang2017deep}, are employed to evaluate the performance of data augmentation by rain generators for deraining. Rain100H is a large-scale dataset, which contains 1800 pairs of rainy and rainless images for training and 100 rainy/rainless images for testing. We also construct two relatively small datasets, Rain100L-S and Rain100H-S, where the training sets are the first 100 pairs of training images from Rain100L and Rain100H, respectively, and the test sets from the test sets of Rain100L and Rain100H, respectively, to further evaluate the performance of data augmentation. We leverage five classical and SOTA deraining models, including PReNet \cite{ren2019progressive}, SPANet \cite{wang2019spatial}, JORDER\_E \cite{yang2019joint}, RCDNet \cite{wang2020model} and DRSformer \cite{chen2023learning}, to evaluate the performance of the proposed generator compared to VRG-Net.

\textbf{Training details of the generator.} For a fair comparison with VRG-Net \cite{wang2021rain}, we exploit the same discriminator as VRG-Net, i.e., a self-attention discriminator \cite{zhang2019self} with gradient penalty. The batch size and patch size are set as 10 and $128\times 128$, respectively. The rotatable TV regularizer is adopted for the training of  Rain100L and Rain100L-S.
{The rain generator is trained for 200 epochs on Rain100L and Rain100L-S, and 400 epochs on Rain100H and Rain100H-S.} The other training settings are those used in Sec. \ref{Dynamism and Controllability}. The augmentation rate is set as 0.5 for Rain100L, Rain100L-S and Rain100H-S, and 1\% for Rain100H. 

\textbf{Evaluation on same-domain datasets.}
Table \ref{Table-syn} provides the deraining results of all competing methods without and with data augmentation, on four datasets. ``Baseline'' denotes the performance of the derainers trained on the original data set without data augmentation here. 
As shown in Table \ref{Table-syn}, the deraining performance of every deep derainer gains a significant improvement with data augmentation in most cases. The performance achieves more improvement on Rain100L-S and Rain100H-S, which accords with human intuition that data augmentation will have more effects when there are less training samples. These results imply that the rainy images generated by the proposed TRG-Net should be closer to those of the original data, which leads to better performance compared to VRG-Net. 

\textbf{Evaluation on cross-domain datasets.}
To further validate the diversity of the samples generated by rain generators in the paired manner, we test the generalization performance of models trained on four augmented datasets. The performance on real rainy images of SPA-Data is shown in Table \ref{Table-spa}. In this case, the PSNR and SSIM with data augmentation by VRG-Net gain an unsubstantial advantage over the baseline in most cases. By comparison, the performance with data augmentation of TRG-Net still achieves a consistently obvious improvement over the baseline in most test cases. These results demonstrate that the samples generated by our method are diverse, which can improve the deraining performance not only in in-distribution but also in out-of-distribution tasks.

\begin{table*}[ht]
\caption{The average PSNR and SSIM of PReNet on the SPA-Data test set. The training set of Baseline is all from original SPA-Data. The training data of VRG-Net and the proposed TRG-Net consist of real pairs randomly selected from SPA-Data and fake pairs generated by different generators. We report the mean of five repeated experiments. The best result is highlighted with \textbf{bold}.}
\label{Table: small_sample}
\centering \setlength{\tabcolsep}{8pt}
{
  \begin{tabular}{c c c c c c c c c c c cc}
    \toprule

 \#Real-samples & \multicolumn{2}{c}{1K} & \multicolumn{2}{c}{1.5K} & \multicolumn{2}{c}{2K} & \multicolumn{2}{c}{3K} & \multicolumn{2}{c}{4K}& \multicolumn{2}{c}{$\sim$630K} \\
 \cmidrule(r){2-3}\cmidrule(r){4-5}\cmidrule(r){6-7}\cmidrule(r){8-9}\cmidrule(r){10-11}\cmidrule(r){12-13}
 &PSNR&SSIM&PSNR&SSIM&PSNR&SSIM&PSNR&SSIM&PSNR&SSIM&PSNR&SSIM \\
 \midrule
 Baseline & 39.41 & 0.9787 & 39.70 & 0.9800 & 39.86 & 0.9809 & 39.96 & 0.9813 & 40.05 & 0.9815 & 40.68 & 0.9845 \\
 \midrule
 \midrule
 \makecell{\#Samples\\(real+fake)} & \multicolumn{2}{c}{1K+0K} & \multicolumn{2}{c}{1K+0.5K} & \multicolumn{2}{c}{1K+1K} & \multicolumn{2}{c}{1K+2K} & \multicolumn{2}{c}{1K+3K} & \multicolumn{2}{c}{2K+2K}\\
 \cmidrule(r){2-3}\cmidrule(r){4-5}\cmidrule(r){6-7}\cmidrule(r){8-9}\cmidrule(r){10-11}\cmidrule(r){12-13}
 &PSNR&SSIM&PSNR&SSIM&PSNR&SSIM&PSNR&SSIM&PSNR&SSIM&PSNR&SSIM \\
 \midrule
 VRG-Net\cite{wang2021rain} & 39.41 & 0.9787 & 39.71 & 0.9796 & 39.83 & 0.9795 & 40.25 & 0.9813 & 40.24 & 0.9814 & 40.73 & 0.9830 \\
 TRG-Net & 39.41 & 0.9787 & \textbf{40.57} & \textbf{0.9826} & \textbf{41.18} & \textbf{0.9839} & \textbf{41.03} & \textbf{0.9834} & \textbf{41.09} & \textbf{0.9834} & \textbf{41.37} & \textbf{0.9850} \\
\bottomrule
\end{tabular}}
\vspace{-0mm}
\end{table*}

\subsubsection{Rain Data Augmentation on Real SPA-Data} \label{Sec: real aug.}

Following \cite{wang2021rain}, we conduct rain data augmentation on SPA-Data \cite{wang2019spatial}, a real dataset containing over 630K pairs of rainy and rainless image patches for training and 1000 pairs of rainy and rainless images for testing, to further verify the diversity and comprehensiveness of the generated rains. 


\textbf{Experiment settings.} 
The generator is trained for 400 epochs. The other training settings of TRG-Net are the same as those settings in Sec. V-C1 of the main text. The rain generators are first trained on SPA-Data in a paired manner. Then we take PReNet \cite{ren2019progressive} as the derainer. The training data for the derainer consist of a small number of pairs (i.e., 1K and 2K) randomly selected from SPA-Data and $N$K fake pairs generated by the generators. For comparison, the same number of paired data from SPA-Data is also randomly chosen as the training set of the derainer, which is called ``Baseline'' in this experiment.

\textbf{The experimental results.} The PSNR and SSIM on the test set of SPA-Data are shown in Table \ref{Table: small_sample}\footnote{To avoid randomness of the experiment, we report the mean of five repetitive experiments following \cite{wang2021rain}.}. From Table \ref{Table: small_sample}, we can observe that better performance can be achieved with generated data samples than that with even more original data samples
In addition, the derainer trained on the samples generated by the proposed TRG-Net obtains a better result than those trained on the same and even more number of data samples generated by the VRG-Net \cite{wang2021rain}. These results validate the superior quality of the generated samples by the proposed TRG-Net beyond those generated by VRG-Net.

\begin{table}
  \caption{The PSNR and SSIM comparison on test data with orientation degree of $[-30^{\circ},30^{\circ}]$ and $\pm[30^{\circ},60^{\circ}]$. Baseline represents the deraining results of PReNet \cite{ren2019progressive} trained on original training set. 
  VRG-Net, TRG-Net$_{[-30^{\circ},30^{\circ}]}$ and TRG-Net$_{[-60^{\circ},60^{\circ}]}$ are the deraining results of PReNet augmented on VRG-Net, TRG-Net$_{[-30^{\circ}\!,30^{\circ}]}$ and TRG-Net$_{[-60^{\circ}\!,60^{\circ}]}$, respectively. 
  }
  \vspace{-2mm}
  \label{Table:Rain100_Gen_theta_v2}

  \centering   \setlength{\tabcolsep}{10pt}
  \begin{tabular}{lcccc}
    \toprule 
    & \multicolumn{2}{c}{in-distribution}  &   \multicolumn{2}{c}{out-of-distribution} \\
    \cmidrule(r){2-3}\cmidrule(r){4-5}
     & \multicolumn{2}{c}{$[-30^{\circ},30^{\circ}]$}  &   \multicolumn{2}{c}{$\pm[30^{\circ},60^{\circ}]$}     \\
    \cmidrule(r){2-3}\cmidrule(r){4-5}
    Methods                       &   PSNR  & SSIM    & PSNR & SSIM  \\
    \midrule
    Input                        &   23.23 & 0.846   & 23.52 & 0.853 \\
    DSC \cite{luo2015removing}   &   24.98 & 0.861   & 24.00 & 0.848 \\
    JCAS \cite{gu2017joint}      &   25.24 & 0.867   & 24.72 & 0.852 \\
    Baseline                     &   37.92 & {0.985}   & 26.58 & 0.885 \\
    VRG-Net \cite{wang2021rain}   &   38.33 & \textbf{0.986}   & 26.52 & 0.886 \\
    TRG-Net$_{[-30^{\circ},30^{\circ}]}$                         &   \textbf{38.58} & \textbf{0.986}   & {27.59} & {0.899} \\
    TRG-Net$_{[-60^{\circ},60^{\circ}]}$           &   {38.44} & \textbf{0.986}   & \textbf{29.06} & \textbf{0.926} \\
    \bottomrule
  \end{tabular}
\end{table}

\subsubsection{OOD Rain Augmentation} \label{Sec: ood aug.}

To further verify the superiority of our controllable rain generation, we generate OOD rain for OOD deraining by controlling the rain factors of trained TRG-Net. Specifically, the rain orientation augmentation are conducted in the following\footnote{We also perform the rain sparsity augmentation experiment in supplementary material.}.

\textbf{Experiment settings.} 
To demonstrate the superiority of our method to control rain orientation, we synthesize 200 pairs of training images containing only -30$^{\circ}$ to 30$^{\circ}$ degree orientation rains from rainless images of Rain100L, and construct two test sets with different orientation degree ranges of rains. One test set contains rains with orientation degrees from -30$^{\circ}$ to 30$^{\circ}$, named as $[-30^{\circ},30^{\circ}]$, and the other contains rains with degrees from -60$^{\circ}$ to -30$^{\circ}$ and 30$^{\circ}$ to 60$^{\circ}$, called $\pm[30^{\circ},60^{\circ}]$. 
In the training phase for deraining networks, we augment the training data with the same orientation as the training set using generators (i.e. VRG-Net and $\mbox{TRG-Net}_{[-30^{\circ},30^{\circ}]}$ in Table \ref{Table:Rain100_Gen_theta_v2}). While for TRG-Net, due to its rain factor controllability, one can augment the training data into a wider range of orientation degrees. Specifically, we augment the orientation degrees of the rains from -60$^{\circ}$ to 60$^{\circ}$ by the proposed TRG-Net, and denote the result as TRG-Net$_{[-60^{\circ},60^{\circ}]}$. 

\textbf{The experimental results.} 
The experimental results are shown in Table \ref{Table:Rain100_Gen_theta_v2}, where ``Baseline'' indicates the deraining result of PReNet trained on original training set.
We also list the deraining results of two model-based methods, DSC \cite{luo2015removing} and JCAS \cite{gu2017joint}.
We can see that Baseline and VRG-Net achieve a relatively poor performance when the orientation of the rain distribution in the test set differs from the training set.
Comparatively, the proposed methods not only perform better on test data with orientation range in $[-30^{\circ},30^{\circ}]$, but also have a well performance on rain orientation range in $\pm[30^{\circ},60^{\circ}]$. Specifically, the augmented training on TRG-Net$_{[-60^{\circ},60^{\circ}]}$ achieves a significant performance improvement in $\pm[30^{\circ},60^{\circ}]$ case.

\begin{table}
  \caption {The FID and KID of generating rain by TRG-Net under different ablation settings.}
  \label{Table:ablation_fid}
  \centering \setlength{\tabcolsep}{10pt}
  \begin{tabular}{ccccc}
    \toprule
    Factor learning & MerNet & RotTV    & FID$\downarrow$  & KID$\downarrow$ \\
    \midrule
    $\times$   & \checkmark & \checkmark    & 42.93 & 0.0131 \\
    \checkmark & $\times$   & \checkmark    & 32.65 & 0.0090 \\
    \checkmark & \checkmark & $\times$      & 29.69 & 0.0073 \\
    \checkmark & \checkmark & \checkmark  & \textbf{25.78} & \textbf{0.0047} \\
    \bottomrule
  \end{tabular} 
\end{table}


\subsection{Ablation Study}

We conduct various ablation settings for the proposed TRG-Net. Firstly, we consider the effectiveness of the rain factor learning. Specifically, we replace transformable convolution kernels with common convolution kernels in our model, which is then degenerated to a CSC-based rain model without any rain factor input. Then we also consider the influence of the merging model and the proposed rotTV regularizer\footnote{For more ablation studies, please refer to the supplementary material.}. All ablation experiments are performed on Rain100L in an unpaired way.
Table \ref{Table:ablation_fid} shows the FID and KID of generating rain by TRG-Net under different ablation settings. 
It can be observed that all the aforementioned settings make a positive contribution to the deraining performance of the entire TRG-Net model, and the completeness of them achieves the best deraining performance.

\section{Conclusion and Future Work}

In this paper, we have constructed a transformable rainy image generation network (TRG-Net), which intrinsically encodes the rain factors, including shape, orientation, length, width and sparsity, into the design of network and can properly extract these fundamental rain factors underlying rainy images purely from data without the need for rain factor labels. Its significance lies in that TRG-Net can not only elaborately design fundamental elements to simulate expected rains, like conventional artificial rendering methods, but also finely adapt to complicated and diverse practical rain patterns, like recent deep learning methods. A transformable convolution framework has been proposed to alleviate the difficulty in embedding controllable and learnable rain factors into deep networks. We have further presented a rotatable TV regularizer for rain generation, which is able to adaptively adjust the orientation for calculating variations and adopt higher penalty along the rain streak orientation. Comprehensive experiments on synthetic and real datasets have validated the superiority of the proposed TRG-Net beyond current SOTA rain generation methods in both unpaired rain generation and paired rain  augmentation tasks.

While the proposed TRG-Net has encoded essential rain factors (e.g., orientation, width, length and density) into the design of the rain generation network in a controllable and interpretable way, other important factors, such as depth and photometry, are simply embedded by the ResNet-based merging network. 
Therefore, the design of the merging network and its integration with modules proposed in previous works should be further explored, especially when applying the proposed method in practical scenarios.
Besides, the proposed transformable convolution framework and the rotatable TV regularizer could have potential application value for extensive tasks, such as the design of the deraining network. We will try these designs in our future research.

\bibliography{TRGNet}

\end{document}